\documentclass[journal]{IEEEtran}
\ifCLASSINFOpdf
\else
\fi
\usepackage{subfiles} % Best loaded last in the preamble
\usepackage{amsmath}
\usepackage{amssymb}
\usepackage{multirow}
\usepackage{chngpage}
\usepackage[english]{babel}
\usepackage{amsthm}
\usepackage{lipsum}
\usepackage{caption}
\usepackage{booktabs}    
\usepackage{graphicx}
\usepackage{hyperref}
\usepackage{enumitem}
\usepackage{subfigure}
\usepackage{csquotes}
\usepackage[dvipsnames]{xcolor}
\usepackage{tikz}
\usepackage{calc} % for simple arithmetic

\theoremstyle{definition}
\newtheorem{definition}{Definition}[section]
\usepackage{xspace}
\newcommand{\latinphrase}[1]{\textit{#1}}\usepackage{xspace}
\newcommand{\etal}{\latinphrase{et~al.}\xspace}

\begin{document}
\title{Graph-based Semi-supervised Learning: A Comprehensive Review}

\author{
    Zixing~Song,
	Xiangli~Yang,
	Zenglin~Xu,~\IEEEmembership{Senior~Member,~IEEE}
	Irwin~King,~\IEEEmembership{Fellow,~IEEE,}
\thanks{Z. Song and I. King are with the Department of Computer Science and Engineering, The Chinese University of Hong Kong, Hong Kong, China(Email: zxsong@cse.cuhk.edu.hk, king@cse.cuhk.edu.hk).}
\thanks{X. Yang is with the SMILE Lab, School of Computer Science and Engineering, University of Electronic Science and Technology of China, Chengdu, China(E-mail: xlyang@std.uestc.edu.cn).}
\thanks{Z. Xu is with  the School of Computer Science and Technology, Harbin Institute of Technology, Shenzhen, China, and also with Peng Cheng Lab, Shenzhen, China (Email: xuzenglin@hit.edu.cn).}
}

\maketitle

\begin{abstract}
Semi-supervised learning (SSL) has tremendous value in practice due to its ability to utilize both labeled data and unlabelled data. An important class of SSL methods is to naturally represent data as graphs such that the label information of unlabelled samples can be inferred from the graphs, which corresponds to graph-based semi-supervised learning (GSSL) methods. GSSL methods have demonstrated their advantages in various domains due to their uniqueness of structure, the universality of applications, and their scalability to large scale data. Focusing on this class of methods, this work aims to provide both researchers and practitioners with a solid and systematic understanding of relevant advances as well as the underlying connections among them. This makes our paper distinct from recent surveys that cover an overall picture of SSL methods while neglecting fundamental understanding of GSSL methods.
In particular, a major contribution of this paper lies in a new generalized taxonomy for GSSL, including graph regularization and graph embedding methods, with the most up-to-date references and useful resources such as codes, datasets, and applications. Furthermore, we present several potential research directions as future work with insights into this rapidly growing field.
\end{abstract}


% \fi
\IEEEpeerreviewmaketitle

\section{Introduction}
\label{sec:introduction}
% with the very first Section~heading (almost always called "Introduction").
% is to refer to the Section~immediately after \Section~in the above as
% \IEEEraisesectionheading puts \Section~within a raised box.

% for IEEE Computer Society journal papers produced under \LaTeX\ using
% IEEEtran.cls version 1.8b and later.

\IEEEPARstart{S}{emi-supervised} learning (SSL) has achieved great successes in various real-world applications where only a few expensive labeled samples are available and abundant unlabeled samples are easily obtained. Moreover, as a typical class of SSL solutions, graph-based SSL (GSSL) is very promising because the graph structure can be naturally used as a reflection for the significant manifold assumption in SSL. More specifically, GSSL methods start with constructing a graph where the nodes represent all the samples and the weighted edges reflect the similarity between a pair of nodes. This way of graph construction implies  that  nodes connected by edge associated with large weights tend to have the same label, which corresponds to the manifold assumption. The manifold assumption suggests that samples locating near to each other on a low-dimensional manifold should share similar labels. Consequently, the expressive power of graph structure under the manifold assumption contributes to the  success of GSSL methods.

In the graph structure commonly used for SSL, each sample is represented by a node, and these nodes are connected by weighted edges that measure the similarity between them. Therefore, the main procedure of GSSL is to create a suitable graph along which the given labels can be easily propagated. To be precise, this goal can be achieved by the following two main steps. 
\begin{enumerate}[label=\textbf{\textit{Step \arabic*.}}, wide=12pt]
\item \textbf{Graph construction.} A similarity graph is constructed based on all the given data, including both the labeled and unlabeled samples. During this step, the biggest challenge is how to make the relationship between original samples well represented. 

\item \textbf{Label inference.} The label inference is performed so that the label information can be propagated from the labeled samples to the unlabeled ones by incorporating the structure information from the constructed graph in the previous step.
\end{enumerate}

Compared with other SSL methods, which are not involved with any graph structure, GSSL has some advantages that are worth noticing. 
In the following, we list several advantages of graph-based SSL methods.
%In summary, the following several reasons are listed to show further why GSSL algorithms are desirable and worth investigating.
\begin{itemize}
    \item \textbf{Universality}. Many common data sets of current trends are represented by graphs like the World Wide Web (WWW), citation networks, and social networks.
    \item \textbf{Convexity}. Since an undirected graph is usually involved in the graph construction step, the symmetric feature of the undirected graph makes it easier to formulate the learning problem into a convex optimization problem, which can be solved with various exciting techniques~\cite{DBLP:series/synthesis/2014Subramanya}.
    \item \textbf{Scalability}. Many of the GSSL methods are meticulously designed so that the time complexity is linear to the total number of samples. As a result, they are often easily parallelized to handle large scale datasets with ease.
\end{itemize}

\begin{figure*}[!t]
	\centering
	\includegraphics[width=0.9\textwidth]{"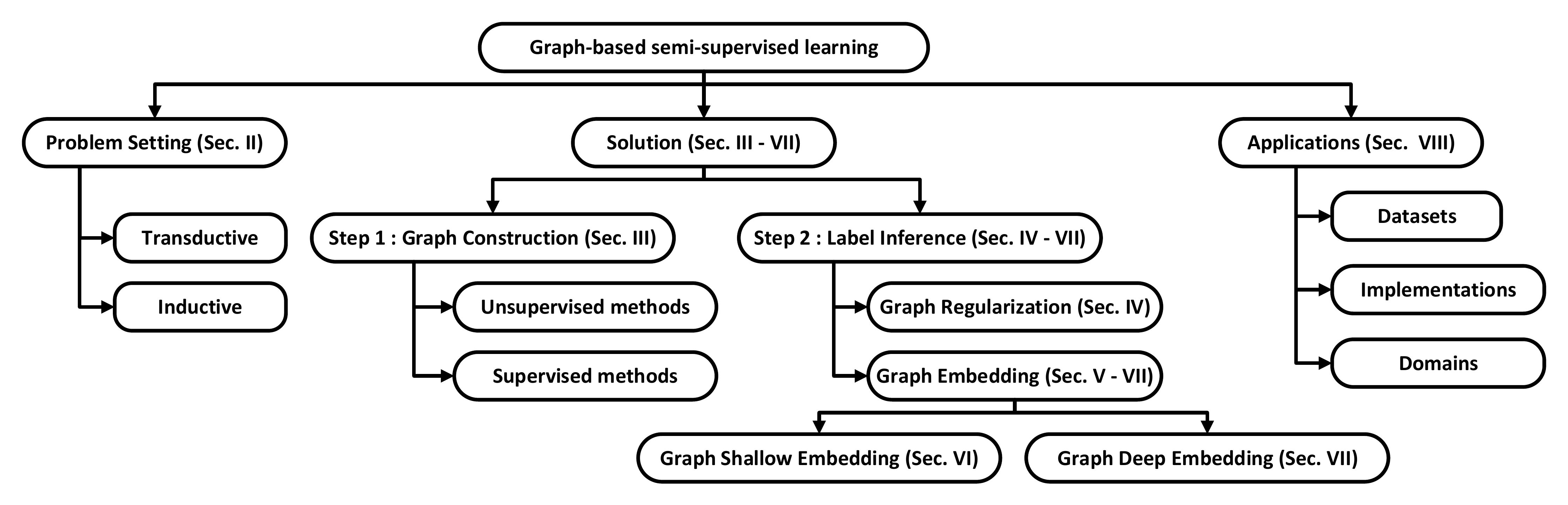"}
	\caption{Taxonomy for graph-based semi-supervised learning}
	\label{gssl_structure}
\end{figure*}

\textbf{Related work.}
Several SSL survey papers~\cite{zhu05survey}\cite{Pise2008SSL} often fail to cover enough methods of GSSL, neglecting its significant role in SSL. Zhu \etal~\cite{zhu05survey} conduct a comprehensive review of classic methods involved in SSL, and GSSL is not explored in detail. Similar earlier work like \cite{Pise2008SSL} by Pise \etal also tries to present a whole picture of SSL methods without covering enough work in GSSL. Recent literature review work,~\cite{PrakashN14}~\cite{van2020survey} and~\cite{ouali2020overview} all follow the footsteps of work~\cite{zhu05survey} and~\cite{Pise2008SSL} by adding more recent research output. However, they do not cover the recent development in GSSL methods. Instead, our work solely focuses on GSSL and combines both earlier studies with recent advances.

The most relevant work to ours is \cite{CHONG2020216} by Chong \etal, and it is considered as the most up-to-date survey paper on GSSL. However, there are several noticeable drawbacks of this work that are worth mentioning here. First, \cite{CHONG2020216} reviews work from the perspective of transductive, inductive, and scalability learning. This taxonomy fails to show the context of development and thus does not reveal the relationship of different methods or models. As a result, we provide a novel taxonomy from the perspective of the two main steps in GSSL: graph construction and label inference. Secondly, some of the reviewed methods in \cite{CHONG2020216} are not graph-based models, but rather are some semi-supervised convolutional neural network (CNN) models as shown in Section 3.4 in the original paper\cite{CHONG2020216}. Most importantly, \cite{CHONG2020216} fails to develop a framework to generalize the methods or models reviewed. However, this paper fills all these gaps with several noticeable contributions.

\textbf{Contributions.} To sum up, this paper presents an extensive and systematic review of GSSL with the following contributions.
\begin{enumerate}
	\item \textbf{Comprehensive review.} We provide the most comprehensive and the most up-to-date overview of GSSL methods. For every approach reviewed in this paper, we present the detailed descriptions with key equations, clarify the context of development beneath the algorithms, make the necessary comparison, and summarise the corresponding strengths or limitations.
	\item \textbf{New taxonomy.} We propose a new taxonomy for graph-based semi-supervised learning with a more generalized framework, as shown in Figure \ref{gssl_structure}. We divide the GSSL process into two steps. The first one is to construct a similarity graph and the second step is to do label inference based on this graph. The latter step is much more challenging and is also the main focus of this paper. Label inference methods are then categorized into two main groups: graph regularization methods and graph embedding methods. For the former group, a generalized framework of regularizers from the perspective of the loss function is presented. For the latter group, we provide a new unified representation for graph embedding methods in SSL with the help of the encoder-decoder framework. 
	\item \textbf{Abundant resources.} We collect abundant resources related to GSSL and build a useful, relevant code base, including the open-source codes for all the reviewed methods or models, some popular benchmark data sets, and pointers to representative practical applications in different areas. This survey can be regarded as a hands-on guide for researchers interested in understanding existing GSSL approaches, using the codes for experiments, and even developing new ideas for GSSL.
	\item \textbf{Future directions.} We propose some open problems and point out some directions for future research in terms of dynamicity, scalability, noise-resilience, and attack-robustness.
\end{enumerate}

\textbf{Organization of the paper.}
The rest of this survey is organized as follows. In Section~\ref{sec:background}, we introduce the background knowledge related to GSSL. Then some necessary notations are listed, and the relevant terms are properly defined. In Section~\ref{gc_section}, we provide a detailed review of graph construction, the first step of GSSL. From Section~\ref{sec:graph regularization} to \ref{sec:Deep Embedding}, the label inference, the second step of GSSL, is covered, which is the main focus of this paper. Furthermore, a new taxonomy is provided, as shown in Figure \ref{gssl_structure}. Graph regularization methods are reviewed in Section~\ref{sec:graph regularization} while graph embedding methods are reviewed from Section~\ref{sec:graph embedding} to Section~\ref{sec:Deep Embedding}. Section~\ref{sec:graph embedding} discusses the generalized encoder-decoder framework for graph embedding. To provide a more detailed overview of it, we further split it into shallow embedding and deep embedding and review them in Section~\ref{shallow_embedding} and Section~\ref{sec:Deep Embedding} respectively. Moreover, in Section~\ref{sec:open_problem}, four open problems are briefly reviewed as future research directions. Finally, applications of GSSL are extensively explored in the Appendix, along with a list of common datasets and a code base for some popular models.

\begin{figure*}[!t]
	\centering
	\includegraphics[width=0.85\textwidth]{"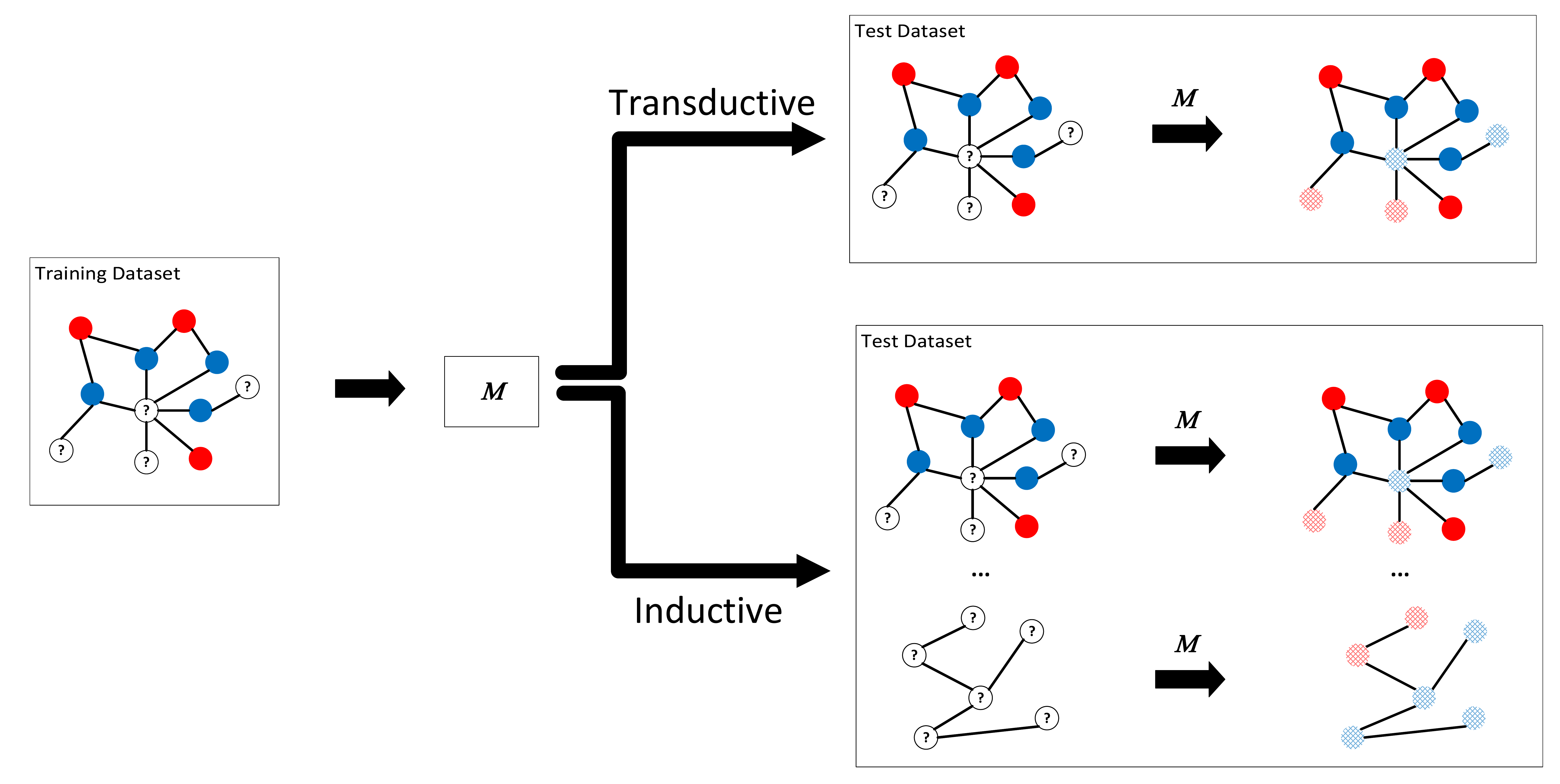"}
	\caption{Comparison between transductive and inductive setting in GSSL. For transductive setting, only the labels of unlabeled nodes in the training dataset need to be inferred while for inductive setting, the trained model $M$ can predict the label of any unseen node.}
	\label{transductive_inductive}
\end{figure*}

\iffalse
\begin{figure*}
  \centering
  \subfigure[Transductive setting]{\includegraphics[scale=0.25]{"./pic/transductive.pdf"}}\quad\quad\quad
  \subfigure[Inductive setting]{\includegraphics[scale=0.25]{"./pic/inductive.pdf"}}
  \caption{Comparison between transductive and inductive setting in GSSL. For transductive setting, only the labels of unlabeled nodes in the training dataset need to be inferred while for inductive setting, the trained model $M$ can predict the label of any unseen node.}
\label{transductive_inductive}
\end{figure*}
\fi
\section{Background and Definition}
\label{sec:background}
As is mentioned earlier, a majority of GSSL algorithms requires solving the following two sub-problems:
\begin{itemize}
	\item Constructing a graph over the input data (if one is not already available).
	\item Inferring the labels on the unlabeled samples in the input or estimating the model parameters. 
\end{itemize}

GSSL methods run on a specifically designed graph in which training samples are represented as nodes, and each node pair is linked by weight to denote the underlying similarity. Some of the nodes are labeled, while others are not. As a result, a graph has to be built to make these problems amenable to the following GSSL approaches.

However, it is worth mentioning that most of the graph-based algorithms are designed for the label inference step. As a result, in this paper, we mainly focus on the label inference techniques used in GSSL, and we only discuss graph construction in Section~\ref{gc_section}. 

Once the graph is constructed, the next step in solving an SSL problem using graph-based methods is to inject labeled data on a subset of the nodes in the graph, followed by inferring the labels for the unlabeled nodes. While a majority of the graph-based inference approaches are transductive, there are some inductive GSSL approaches as well. 

Following the framework for SSL, the loss function of GSSL approaches can also be generalized within that of SSL, which contains three parts as shown in Eq.~(\ref{general_ssl_loss})
\begin{equation}
	\label{general_ssl_loss}
	\mathcal{L}(f) = \underbrace{\mathcal{L}_{s}(f,\mathcal{D}_l)}_{\text{supervised loss}} + \lambda \underbrace{\mathcal{L}_{u}(f,\mathcal{D}_u)}_{\text{unsupervised loss}} + \mu \underbrace{\mathcal{L}_r(f,\mathcal{{D}})}_{\text{regularization loss}},
\end{equation}
where $\mathcal{L}_{s}(f,\mathcal{D}_l)$ is the supervised loss on the labeled data and $\mathcal{L}_{u}(f,\mathcal{D}_u)$ is the unsupervised loss on the unlabeled data and $\mathcal{L}_r(f,\mathcal{{D}})$ is the regularization loss. Additionally, $\lambda$ and $\mu$ are hyperparameters to balance these terms. However, for GSSL, unsupervised loss is often absorbed into the regularization loss since no label information is used in the regularization loss term. Therefore, the loss function for GSSL can be generalized as shown in Eq.~(\ref{general_gssl_loss})  
\begin{equation}
	\label{general_gssl_loss}
	\mathcal{L}(f) = \mathcal{L}_{s}(f,\mathcal{D}_l) + \mu \mathcal{L}_r(f,\mathcal{{D}}).
\end{equation}

In this paper, more attention will be paid to how to do label inference when the similarity graph has already been constructed from the given datasets under the setting of semi-supervised learning. Two main groups of GSSL are reviewed following Eq.~(\ref{general_gssl_loss}): graph regularization and graph embedding. More details will be provided in Section~\ref{sec:graph regularization} to Section~\ref{sec:Deep Embedding}.

\subsection{Related concepts}
\subsubsection{{Supervised learning and unsupervised learning}}
Supervised learning and unsupervised learning can be viewed as two extremes of SSL because all the training samples are well labeled in supervised learning settings, while unsupervised learning can only have access to unlabeled data. Semi-supervised learning aims to introduce cheap unlabeled samples to enhance the model's performance with only a few costly labeled samples. Therefore, the problem setting of SSL is a perfect match for many real-world applications.
\subsubsection{Other semi-supervised learning methods}
Throughout the development of SSL, a great number of successful algorithms or models have emerged in roughly three phases. The first phase is the early stage of SSL before 2000, where classic machine learning algorithms are investigated and improved with unlabeled data. Typical examples are S3VM and Co-training. The second phase is the mature stage of SSL between 2000 and 2015, in which many methods flourished, such as mixture model, pseudo label, self-training, manifold learning, and GSSL. The third phase is after 2015, with the advance of deep learning and especially Graph Neural Networks (GNN). Since GSSL methods witness all these three stages, reviewing its development and recent progress is necessary.

\subsubsection{{Transductive and inductive settings}}
Like other SSL methods, GSSL algorithms can be divided into two categories based on whether to predict data samples' labels out of training data.

\begin{definition}[\textbf{Transductive setting}]
Given a training set consisting of labeled and unlabeled data $\mathcal{D}=\left\{\left\{\mathbf{x}_{i}, y_{i}\right\}_{i=1}^{n_{l}},\left\{\mathbf{x}_{i}\right\}_{i=1}^{n_{u}}\right\}$, the goal of a transductive algorithm is to learn a function $f : \mathcal{X} \rightarrow \mathcal{Y}$ so that $f$ is only able to predict the labels for the unlabeled data $\left\{\mathbf{x}_{i}\right\}_{i=1}^{n_{u}}$.
\end{definition}
 
\begin{definition}[\textbf{Inductive setting}]
Given a training set consisting of labeled and unlabeled data $\mathcal{D}=\left\{\left\{\mathbf{x}_{i}, y_{i}\right\}_{i=1}^{n_{l}},\left\{\mathbf{x}_{i}\right\}_{i=1}^{n_{u}}\right\}$, the goal of an inductive algorithm is to learn a function $f : \mathcal{X} \rightarrow \mathcal{Y}$ so that $f$ is able to predict the output $y$ of any input $x \in \mathcal{X}$.
\end{definition}

While most of the GSSL approaches are transductive, there are a few inductive GSSL approaches. In most scenarios, transductive SSL methods outperform inductive ones in terms of prediction accuracy while they often suffer from high training costs compared to inductive ones, especially in the context of large scale incremental learning. Figure \ref{transductive_inductive} illustrates the difference between transductive and inductive setting in GSSL.

\subsection{Notations and Definitions}
In this section, as a matter of convenience, we first define some useful and common terms used in GSSL, along with relevant notations. Unless otherwise specified, the notations used in this survey paper are illustrated as Table \ref{notations}. After the list of the notations, the minimal set of definitions required to understand this paper is defined.

\begin{table}[!t]
	\centering
	\caption{Notations used in the paper}
	\label{notations}
%	\resizebox{\textwidth}{!}{%
		\begin{tabular}{cc}
			\toprule
			\textbf{Notations}                                       & \textbf{Descriptions}                                \\ \midrule
			$G$                                                      & A Graph                                              \\
			$V$                                                      & The set of nodes (vertices) in a graph               \\
			$E$                                                      & The set of edges in a graph                          \\
			$i$,$v$                                                  & Node $i$, Node $v$                                   \\
			$(i,j)$                                                  & The edge linked between node $i$, $j$        \\
			$W$                                                      & The weight matrix of a graph            \\
			$W_{ij}$                                                & The weight associated with edge $(i,j)$ \\
			$A$                                                      & The adjacency matrix of a graph                      \\
			$A_{ij}$                                       &$i^{th}$ row $j^{th}$ column in the adjacency matrix $A$\\
			$D$                                                      & The degree matrix of a graph                         \\
			$D_{ii}$                                                 & The degree of node $i$                               \\
			$X$                                                      & The attribute matrix of a graph                       \\
			$x_{i}$                                                  & The attribute vector for node $i$ \\
			$\mathcal{N}(i)$                                         & The neighborhood of a node $i$                       \\
			$L$                                                      & Unnormalized graph Laplacian matrix                  \\
			$\tilde{L}$                                            & Normalized graph Laplacian matrix                    \\
			$\mathcal{P}^{1}_{i}$                                    & First-order proximity of node $i$                    \\
			$\mathcal{P}^{2}_{i,j}$                                  & Second-order proximity between node $i$, $j$ \\
			$n_l$                                                    & The number of labeled samples                        \\
			$n_u$                                                    & The number of unlabeled samples                      \\
			$\mathcal{D}_l=\{\mathbf{x}_{i}, y_{i}\}_{i=1}^{n_{l}}$ & Labeled samples                                      \\
			$\mathcal{D}_u=\{\mathbf{x}_{i}\}_{i=1}^{n_{u}}$        & Unlabeled samples                                    \\
			$\mathbf{S}$                                            & Similarity matrix of a graph                 \\   
			$\mathbf{S}[u,v]$                                       & Similarity measurement between node $u$,$v$ \\    
			$\mathbf{Z}$                                   		& Embedding matrix                                \\
			$\mathbf{z}_{i}$                                   		& Embedding for node $i$                                \\
			$\mathbf{h}_{v}^{(k)}$                                    & Hidden embedding for node $v$ in $k^{th}$layer \\           
			$\mathbf{m}_{\mathcal{N}(v)}^{(k)}$		                & \begin{tabular}[x]{@{}c@{}}Message aggregated from node $v$'s \\neighborhood in $k^{th}$layer\end{tabular} \\ \bottomrule
				\end{tabular}%
%			}
\end{table}

\begin{definition}[\textbf{Graph}]
	\label{Graph}
A graph is an ordered pair $G=(V,E)$ where $V=\{1, \ldots,|V|\}$ is the set of nodes (or vertices) and $E \subseteq\{V \times V\}$ is the set of edges.

GSSL algorithms start by representing the data as a graph. We assume that the node $i \in V$ represents the input sample $\mathbf{x}_{i}$. We will be using both $i$ and $\mathbf{x}_{i}$ to refer to the $i^{th}$ node in the graph.
\end{definition}

\begin{definition}[\textbf{Directed and Undirected Graphs}]	
A graph whose edges have no starting or ending nodes is called an undirected graph. In the case of a directed graph, the edges have a direction associated with them. 
\end{definition}

\begin{definition}[\textbf{Weighted Graph}]
A graph $G$ is weighted if there is a number or weight associated with every edge in the graph. Given an edge $(i,j)$, where $i, j \in V$ and $W_{i j}$ is used to denote the weight associated with the edge $(i,j)$ and thus forms the whole weight matrix $W \in \mathcal{R}^{n \times n}$. In most cases, we assume $W_{i j} \geq 0$ and $W_{i j}$ can be $0$ if and only if there is no edge between the node pair $(i,j)$. 
\end{definition}

\begin{definition}[\textbf{Degree of a Node}]
The degree $D_{ii}$ of the node $i$ is given by  $D_{i i}=\sum_{j} W_{i j}$. Moreover, in the case of an unweighted graph, the node's degree is equal to its number of neighbors.
\end{definition}

\begin{definition}[\textbf{Neighborhood of a Node}]
The neighborhood of a node $v$ in a graph $G$ is denoted as $\mathcal{N}(v)$ to indicate the subgraph of $G$ induced by all nodes adjacent to $v$.
\end{definition}

\begin{definition}[\textbf{Adjacency Matrix}]
	\label{Adjacency_Matrix}
	 Adjacency matrix is a matrix with a 1 or 0 in each position $(i,j)$ based on whether node $i$ and node $j$ are adjacent or not. If the given graph is undirected, its corresponding adjacency matrix is a symmetric matrix.
\end{definition}

\begin{definition}[\textbf{Graph Laplacian Matrix}]
	\label{Laplacian_Matrix}
The unnormalized graph Laplacian matrix is given by $L=D-W$. Here the $D \in \mathcal{R}^{n \times n}$ is a diagonal matrix such that $D_{i,i}$ is the degree of the node $i$ and otherwise $D_{i j}=0$ $\forall i \neq j$. It is easy to prove that $L$ is a  positive semi-definite matrix.

The normalized graph Laplacian matrix is given by $\tilde{L}=D^{-1 / 2} L D^{1 / 2}$ where $L$ is the unnormalized graph Laplacian matrix.
\end{definition}

\section{Graph Construction}
\label{gc_section}
To perform any GSSL methods, a graph must be constructed first, where nodes represent data samples, some of which are labeled while others are not, and edges are associated with a certain weight to reflect each node pair's similarity. In some domains, such as citation networks, there is already an implicit underlying graph. Graph-based methods are thus a natural fit for SSL problems in these domains. For most of the other machine learning tasks, however, it is believed that the data instances are not conveniently represented as a graph structure, and as a result, a graph has to be built to make these problems appropriate for GSSL approaches. The graph construction techniques are involved in the first step mentioned before.

% \subsection{Problem statement}
% Recall that the training data in the case of SSL is $\mathcal{D}=\left\{\mathcal{D}_{l}, \mathcal{D}_{u}\right\}$, where $\mathcal{D}_{l}=\left\{\left(\mathbf{x}_{i}, y_{i}\right)\right\}_{i=1}^{n_{l}}$ are the labeled samples and $\mathcal{D}_{u}=\left\{\mathbf{x}_{i}\right\}_{i=1}^{n_{u}}$ are the unlabeled samples. And $n = n_{l} +n_{u}$ is the total number of the training samples.
The goal of graph construction is to discover a graph $G=(V, E, W)$ where $V$ is the set of nodes, $E \subseteq V \times V$ are the edges, and $W \in \mathcal{R}^{n \times n}$ are the associated weights on the edges. Each node in the graph represents an input sample, and thus the number of nodes in the graph is $|V|=n$. As the nodes are fixed (assuming that $D$ is fixed, which is often the case), the task of graph construction involves estimating $E$ and $W$. The following three assumptions often hold.
\begin{enumerate}[label=\textit{Assumption \arabic*.}, wide=12pt]
\item The graph is undirected, so $W$ is symmetric. And all edge weights
are non-negative, $W_{ij} \geq 0$, $\forall i \neq j$.
\item $W_{ij} = 0$ implies the absence of an edge between nodes $i$ and $j$.
\item There are no self-loops, $W_{ii} = 0$, $\forall 1\leq i \leq n$.
\end{enumerate}
These three assumptions simplify the problem by adding these constraints and lay the foundations for the following unsupervised and supervised methods.
\subsection{Unsupervised methods}
Unsupervised graph construction techniques ignore all the given label information of the training data during the construction process. Among all the unsupervised methods for graph construction, the K-nearest neighbor (KNN) graph and b-Matching methods, along with their extensions, are the most popular ones.

\subsubsection{KNN-based approaches}
\label{sec:knn}
For KNN-based graph construction approaches~\cite{sslbook}, every node is associated based on a pre-configured distance metric with its $k$ nearest neighbors in the resulting graph. Moreover, KNN-based methods link the $k$ nearest neighbors greedily to generate graphs whose nodes' degree is larger than $k$, which leads to irregular graphs. Note that a graph is said to be regular if every node has the same degree.

KNN-based method needs a proximity function $\texttt{sim}\left(\mathbf{x}_{i}, \mathbf{x}_{j}\right)$ or distance metric that can quantify the resemblance or disparity between every node pair in the training data. The weight value associated with the edge is given by Eq.~(\ref{knn_sim}),
\begin{equation}
	\label{knn_sim}
	W_{ij}=
\left\{
\begin{array}{ll}
	\texttt{sim}\left(\mathbf{x}_{i}, \mathbf{x}_{j}\right) & i \in \mathcal{N}(j)\\
	0 & otherwise \\
\end{array}. 
\right.
\end{equation}

In $\varepsilon$-neighborhood-based graph construction method~\cite{sslbook}, if the distance between a node pair is smaller than $\varepsilon$, where $\varepsilon \geq 0$ is a predefined constant, a connected edge is formed between them. KNN methods enjoy certain favorable properties when compared with $\varepsilon$-neighborhood-based graphs. Specifically, in $\varepsilon$-neighborhood-based graphs, a misleading choice of the parameter $\varepsilon$ could lead to generating disconnected graphs~\cite{bmatching}. However, KNN-based graphs outperform $\varepsilon$-neighborhood-based ones with better scalability.

In Oziki~\etal~\cite{mKNN}, it is contended that a hub or a center situated in the sample space can result in a corresponding hub in the classic KNN graphs. This may downgrade the prediction performance on several classification tasks. In order to handle this issue, ~\cite{mKNN} proposes a new way of constructing a graph by using mutual KNN in combination with a maximum span tree (M-KNN). In parallel with this work, Vega~\etal~\cite{vega2014regular} also introduce the sequential KNN (S-KNN) to produce graphs under the new relaxed condition in which the resulting graph contains no hubs but is not necessarily regular.

\subsubsection{b-Matching}
As discussed above, KNN graphs, contrary to their name, often lead to graphs where different nodes have different degrees. Jebara \etal~\cite{bmatching} propose b-Matching, which guarantees that every node in the resulting graph has exactly $b$ neighbors. Using b-Matching for graph construction involves two steps: (a) graph sparsification and (b) edge re-weighting.

In graph sparsification, there exists an issue where edges are removed in a way of estimating a matrix $P \in\{0,1\}^{n \times n}$. For the entry in $P$, $P_{i j}=1$ signifies an existing edge between a node pair in the generated graph, while $P_{i j}=0$ suggests a lack of an edge. b-Matching provides a solution by formulating an optimization problem with the objective,
\begin{equation}
	\begin{gathered}
		\min _{P \in\{0,1\}^{n \times n}} \sum_{i, j} P_{i j} \Delta_{i j} \\
		s.t. \sum_{j} P_{i j}=b, P_{i i}=0, P_{i j}=P_{j i}, \forall 1 \leq i, j \leq n.
	\end{gathered}
\end{equation}
Here, $\Delta \in \mathcal{R}_{+}^{n \times n}$ is a symmetric distance matrix. 

When a selection of edges is made in matrix $P$ from the previous step, the next aim is to determine the chosen edges' associated weights to produce the estimated weight matrix $W$. There are three popular ways to determine the weight matrix $W$.

\begin{itemize}
    \item \textbf{Binary Kernel.} The easiest way to estimate $W$ is to set $W = P$. Thus $W_{ij} = P_{ij}$ and each entity in $W$ is also either 0 or 1.
    \item \textbf{Gaussian Kernel.} Here, $W$ can be a little bit complex compared to the previous one. That is
    $W_{i j}=P_{i j} \exp \left(-\frac{d\left({x}_{i}, {x}_{j}\right)}{2 \sigma^{2}}\right)$.
    \item \textbf{Locally Linear Reconstruction (LLR).} LLR is derived from the Locally Linear Embedding (LLE) technique by Roweis \etal~\cite{Roweis2323}. The goal is to reconstruct $\mathbf{X}_{i}$ from its neighborhood. It can be formulated to solve the following optimization problem,
    \begin{equation}
    	\begin{gathered}
    		\min _{W} \sum_{i}\left\|{x}_{i}-\sum_{j} P_{i j} W_{i j} {x}_{j}\right\|^{2}\\ s.t. \sum_{j} W_{i j}=1, W_{i j} \geq 0, i=1, \ldots, n.
    	\end{gathered}
    \end{equation}
\end{itemize}

In summary, the b-matching method restricts the constructed similarity graph to be regular so that the given label can be propagated in a more balanced way during the following label inference step.

\iffalse
\textcolor{red}{needs paraphrasing}
\textbf{Summary}

% \begin{table}[]
% \begin{tabular}{cccc}
% \hline
% kNN         & S-kNN       & M-kNN          & b-Matching  \\ \hline
% $O(dn^{2})$ & $O(dn^{2})$ & $O(n^{2}logn)$ & $O(bn^{3})$ \\ \hline
% \end{tabular}
% \end{table}

\begin{table}[!t]
\renewcommand{\arraystretch}{1.3}
\caption{Computational Cost of Unsupervised Methods for Graph Construction}
\label{time_complexity}
\centering
\begin{tabular}{cccc}
\hline
KNN         & S-KNN       & M-KNN          & b-Matching  \\ \hline
$O(dn^{2})$ & $O(dn^{2})$ & $O(n^{2}+ kn\log n)$ & $O(bn^{3})$ \\ \hline
\end{tabular}
\end{table}

In Table \ref{time_complexity}, we summarize the time complexity of the mentioned unsupervised methods for graph construction. Both KNN and S-KNN are well-known. For M-KNN, We first compute a minimum spanning tree from the complete weighted graph, which takes the complexity $O(n^{2}\log n)$. Ozaki \etal~\cite{mKNN} use a Fibonacci heap-based implementation and constructed the mutual KNN graph in $O(n^{2}+ kn\log n)$. Huang \etal~\cite{bmatching-impl} show that the running time of the implementation for b-Matching is $O(bn^{3})$.

According to Berton \etal~\cite{graph_construction_compare}, the experimental results indicate that regular graphs, generated by b-Matching and S-KNN, achieve the best results on classification and have a lower average degree and homogeneous topology. However, S-KNN is less time consuming for execution compared with b-Matching.
\fi

\subsection{Supervised methods}
\iffalse
\textcolor{red}{may add some other models and add formula to existing models}
\fi
The existing prevalent strategies of graph construction are unsupervised, i.e., they fail to use any given label information during the construction phase. However, labeled samples can be used as a kind of prior knowledge that can be used to refine the generated graph for the downstream learning tasks. Dhillon \etal~\cite{dhillon2010inference} study the possibility of employing labeled points so as to measure the similarities between node pairs. Rohban \etal~\cite{ROHBAN20121363} suggest another supervised method of graph construction, which demonstrates that the optimal solution for a neighborhood graph can be regarded as a subgraph of a KNN graph as long as the manifold sampling rate is large enough.

Driven by previous studies~\cite{mKNN}, a new method, graph-based on informativeness of
labeled instances (GBILI)~\cite{GBILI}, also utilizing the label information, is introduced. GBILI not only results in a decent accuracy on classification tasks but also stands out with a quadratic time complexity~\cite{berton2015graph}. Moreover, built on GBILI~\cite{GBILI}, Lilian \etal~\cite{RGCLI} have upgraded the method for producing more robust graphs by solving an optimization problem with the specific algorithm called the Robust Graph that Considers Labeled Instances (RGCLI). More recently, a new SSL learning method referred to as a low-rank semi-supervised representation is proposed~\cite{7931596} which incorporates labeled data into the low-rank representation (LRR). A follow-up work is by Taherkhani~\etal~\cite{taherkhani2019matrix}. By taking additional supervised information, the generated similarity graph can facilitate the following label inference process to a great extent.

\section{Graph regularization}
\label{sec:graph regularization}
All the classic GSSL methods can actually be simplified as searching for a function $f$ on the graph. $f$ has to satisfy two criteria simultaneously: 1) it must be as close to the given labels as possible, and 2) it must be smooth on the entire constructed graph. 

These two conditions can be further expressed in a general regularization framework in which loss function can be decomposed into two main parts. The first term is a supervised loss constraint to the first criterion, and the second term is a graph regularization loss constraint to the second criterion. Formally, we have,
\begin{equation}
\label{general_graph_regularization}
    \mathcal{L}(f) = \sum_{(x_{i},y_{i}) \in \mathcal{D}_{l}}\underbrace{\mathcal{L}_{s}(f(x_{i}),y_{i})}_{\text{supervised loss}} + \mu \sum_{x_i \in \mathcal{D}_l+\mathcal{D}_u}\underbrace{\mathcal{L}_{r}(f(x_{i}))}_{\text{regularization loss}},
\end{equation}
where $f$ is the prediction function and $\mu$ is a trade-off hyper-parameter. 

In the following sections, we will see that all the graph regularization methods reviewed here are similar. They only differ in the particular choice of the loss function with various regularizers. 
% Figure \ref{outline_graph_regularization} provides the outline of the reviewed approaches in Section~\ref{sec:graph regularization}. 
Table \ref{tab:summary_graph_regularization} summarizes all the reviewed graph regularization methods in Section~\ref{sec:graph regularization} from the perspective of decomposing the regularizer. This generalized framework of graph regularization has been carefully examined by Zhou \etal~\cite{zhou2004regularization}, and its theoretical analysis from different perspectives has also been provided by~\cite{ando2007learning} ~\cite{calder2019properly}~\cite{hoffmann2020consistency}. 
\iffalse
\begin{figure}[!t]
	\centering
	\includegraphics[width=0.37\textwidth]{"./pic/graph_regularization_methods_in_GSSL.pdf"}
	\caption{Graph Regularization methods in GSSL}
	\label{outline_graph_regularization}
\end{figure}
\fi
\begin{table*}[!t]
\centering
\caption{Summary on Graph Regularization Methods}
\label{tab:summary_graph_regularization}
\begin{tabular}{ccc}
\toprule
\textbf{Method} &
  \textbf{Supervised loss} $f_{s}(f,\mathcal{D}_l)$ &
  \textbf{Graph regularization loss} $f_{r}(\mathcal{D})$ \\
  \midrule
GRF~\cite{zhu2003semi} &
  $\sum_{i=1}^{n_l}\left(f({x_i})-y_{i}\right)^{2}$ &
  $\sum_{i, j} W_{i j}\left(f(x_i)-f(x_j)\right)^{2}$ \\
LRC~\cite{zhou2004learning} &
  $\sum_{i=1}^{n_l}\left(f({x_i})-y_{i}\right)^{2}$ &
  $\left(\sum_{i,j} W_{i j}\left(\frac{1}{\sqrt{D_{i i}}} f({x_i})-\frac{1}{\sqrt{D_{j j}}} f({x_j})\right)^{2}\right)$ \\
p-Laplacian~\cite{slepcev2019analysis} &
  $\sum_{i=1}^{n_l}\left(f({x_i})-y_{i}\right)^{2}$ &
  $\sum_{i,j} W_{i j}\left|\frac{1}{\sqrt{D_{i i}}} f({x_i})-\frac{1}{\sqrt{D_{j j}}} f({x_j})\right|^{p}$\\
Directed regularization~\cite{zhou2005icml} &
  $\sum_{i=1}^{n_l}\left(f({x_i})-y_{i}\right)^{2}$ &
  $\left(\sum_{i,j} \pi(i)p(i,j) \left(\frac{1}{\sqrt{D_{i i}}} f({x_i})-\frac{1}{\sqrt{D_{j j}}} f({x_j})\right)^{2}\right)$\\
Manifold regularization~\cite{DBLP:journals/jmlr/BelkinNS06} &
  $\sum_{i=1}^{n_l}\left(f(x_i) - y_i\right)^2$ &
  $\gamma_A \|f\|_{K}^{2} + \gamma_I \frac{1}{(n_l + n_u)^2} \hat{y}^T L \hat{y}$\\
LPDGL~\cite{7010929} &
  $\sum_{i=1}^{n_l}\left(f({x_i})-y_{i}\right)^{2}$ &
  $\left(\sum_{i,j} W_{ij}\left(f(x_i) - f(x_j)\right)^{2}\right) + \left(\sum_{i=1}^{n} (1 - \frac{D_{ii}}{\sum_{j=1}^{n}D_{jj}})\left(f(x_i)\right)^2\right)$ \\
Poisson learning~\cite{calder2020poisson} &
  $\sum_{i=1}^{n_l}\left(f({x_i})-y_{i}\right)^{2}$ &
  $\left(\sum_{i,j} W_{ij}\left(f(x_i) - \sum_{j \in \mathcal{N}(i)}f(x_j)\right)^{2}\right)$  \\
  \bottomrule
  
\end{tabular}
\end{table*}

\subsection{Label propagation}
\label{sec:Label propagation}
Label Propagation (LP)~\cite{zhuLP} is the most popular method for label inference on GSSL. Label Propagation can be formulated as a problem, in which some of the nodes' labels, also referred to as seeds, propagate to unlabeled nodes based on the similarity of each node pair, which is represented by the constructed graph discussed in Section~\ref{gc_section}. Meanwhile, during the propagation process, given labels need to be fixed. In this way, labeled nodes serve as guides that lead label information flow through the edges within the graph so that unlabeled nodes can also be tagged with predicted labels.

The basic version of label propagation algorithm is as follows:
\begin{enumerate}[label=\textit{Step \arabic*.}, wide=12pt]
    \item All nodes propagate labels for one step $Y \leftarrow T Y$.
    \item Row-normalize $Y$ to maintain the class probability interpretation.
    \item Clamp the labeled data. Repeat from step 2 until $Y$ converges.
\end{enumerate}

\subsubsection{Gaussian random fields}
Gaussian Random Fields (GRF)~\cite{zhu2003semi} is a typical example of the early work in GSSL by using label propagation algorithms. The strategy is to estimate some prediction function $f$ based on the graph $G$ with some constraints to ensure certain necessary properties and afterward attach labels to the unlabeled nodes according to $f$.  In fact, the above-mentioned constraint is to take $f(x_i)=f_{l}(x_i) \equiv y_{i}$ on all the labeled nodes. Intuitively, the clustering unlabeled points with strongly connected edges should share common labels. This is why the quadratic energy function is designed as shown in Eq.~(\ref{quadratic energy function}),

\begin{equation}
	\label{quadratic energy function}
    E(f)= \mathcal{L}_{r}= \frac{1}{2} \sum_{i, j} W_{i j}\left(f(x_i)-f(x_j)\right)^{2}.
\end{equation}
It is noteworthy that the minimum value of energy function $f=\arg \min _{\left.f\right|_{\mathcal{D}_l}=f_{l}} E(f)$ is harmonic; namely, it satisfies the constraint $L f=0$ on the unlabeled nodes and is equal to $f_{l}$ on the labeled nodes $\mathcal{D}_l$, where $L$ is the graph Laplacian matrix.

The property of harmonic function indicates that the value of $f$ at every unlabeled node is the mean value of $f$ at its neighboring nodes: $f(x_j)=\frac{1}{d_{j}} \sum_{i \sim j} W_{i j} f(x_i),$ for $j=l+1, \ldots, l+u$. This constraint is actually compatible with the previous smoothness requirement of $f$ with respect to the graph. It can also be interpreted in an iterative manner as shown in Eq.~(\ref{1.1.1})
\begin{equation}
    f^{(t+1)} \leftarrow P \cdot f^{(t)},
    \label{1.1.1}
\end{equation}
where $P=D^{-1} W$.
Furthermore, a closed form solution of Eq.~(\ref{1.1.1}) can be deduced if weight matrix $W$ is split into four blocks $W=\left[\begin{array}{ll}W_{l l} & W_{l u} \\ W_{u l} & W_{u u}\end{array}\right]$. Then, 
\begin{equation}
    f_{u}=\left(D_{u u}-W_{u u}\right)^{-1} W_{u l} f_{l}=\left(I-P_{u u}\right)^{-1} P_{u l} f_{l}.
\end{equation}
\iffalse
The use of a Gaussian kernel to construct the graph in the original paper~\cite{zhu2003semi} is the source of the word “Gaussian” in GRF. However, we note that GRF is more of an inference algorithm and can be viewed as an example of LP method. In fact, it can be applied to arbitrary graphs, and not necessarily restricted to those constructed using Gaussian kernels alone.
\fi
\subsubsection{Local and global consistency}
Zhou \etal~\cite{zhou2004learning} extend the work~\cite{zhu2003semi} to multiclass setting and proposes Local and Global Consistency (LGC) to handle a more general semi-supervised problem. The iterative formula is shown in Eq.~(\ref{1.1.2})
\begin{equation}
    Y^{(t)}=\alpha \mathcal{S} Y^{(t-1)}+(1-\alpha) Y^{(0)},
    \label{1.1.2}
\end{equation}
where $\mathcal{S}=D^{-1 / 2} A D^{-1 / 2}$, and $\alpha$ is a hyper-parameter. We can also easily derive the closed-form solution for Eq.~(\ref{1.1.2}) as shown in Eq.~(\ref{1.1.3})
\begin{equation}\hat{Y}=\alpha \mathcal{S} \hat{Y}+(1-\alpha) Y^{(0)}. \label{1.1.3}
\end{equation}
From a perspective of optimization problem, LGC~\cite{zhou2004learning} actually tries to minimize the following objective function Eq.~(\ref{1.1.2.1}) associated with prediction function $f$.
% \begin{equation}
\begin{align}	
	\mathcal{L}(f)=\frac{1}{2}&\left(\sum_{i,j} W_{i j}\left(\frac{1}{\sqrt{D_{i i}}} f({x_i})-\frac{1}{\sqrt{D_{j j}}} f({x_j})\right)^{2}\right)+ \nonumber \\ 
	&\mu \sum_{i=1}^{n_l}\left(f({x_i})-y_{i}\right)^{2}.\label{1.1.2.1}
\end{align}
% \end{equation}

Compared to the GRF objective above, LGC has two important differences: (a) the inferred labels for the ``labeled`` nodes are no longer required to be exactly equal to the seed values, and this helps with cases where there may be noise in the seed labels, and (b) the label for each node is penalized by the degree of that node $\frac{1}{\sqrt{D_{i i}}}$, ensuring that in the case of irregular graphs, the influence of high degree nodes is regularized.

There exist quite a few variants of LGC method, a representative one is p-Laplacian regularization~\cite{slepcev2019analysis}. The first term in Eq.~(\ref{1.1.2.1}) can be substituted by a more general one as $\sum_{i,j} W_{i j}\left|\frac{1}{\sqrt{D_{i i}}} f({x_i})-\frac{1}{\sqrt{D_{j j}}} f({x_j})\right|^{p}$, where $p$ is a positive integer. Slepcev \etal~\cite{slepcev2019analysis} provide a comprehensive analysis of its theoretical grounds. In addition, many applications based on LGC are proven to be successful in various domains. For example, Iscen \etal~\cite{iscen2019label} utilize LGC method to facilitate the training process of deep neural networks (DNNs) by generating pseudo-label for the unlabeled data.

\subsection{Directed regularization}
In the previous label propagation methods, only undirected graphs are applicable. A new regularization framework for directed graphs, such as citation networks, is provided to solve this issue~\cite{zhou2005icml}. To fully take the directionality of the edges into consideration, the idea of naive random walk is incorporated into this regularization framework. $\pi$ is used to denote a unique probability distribution satisfying the following equations,
\begin{equation}
    \label{balance_eq}
    \pi(i) = \sum_{j \rightarrow i}{\pi(j)p(j,i)}, \forall i \in V,
\end{equation}
where
\begin{equation}
\label{def_pij}
    p(i,j) = \frac{W_{ij}}{d^+ (i)} = \frac{W_{ij}}{\sum_{j \leftarrow i}{W_{ij}}}.
\end{equation}
In Eq.~(\ref{balance_eq}) and Eq.~(\ref{def_pij}), $j \rightarrow i$ denotes the set of vertices adjacent to the vertex $i$ while $j \leftarrow i$ denotes the set of vertices adjacent from the vertex $i$. Thus, we can define a loss function that sums the weighted variation of each edge in the directed graph as shown in Eq.~(\ref{directed_eq}).
\begin{align}	
	\mathcal{L}(f)=\frac{1}{2}&\left(\sum_{i,j} \pi(i)p(i,j) \left(\frac{1}{\sqrt{D_{i i}}} f({x_i})-\frac{1}{\sqrt{D_{j j}}} f({x_j})\right)^{2}\right)\nonumber \\ 
	&+ \mu \sum_{i=1}^{n_l}\left(f({i})-y_{i}\right)^{2}. \label{directed_eq}
\end{align}

It is also worth noting that Eq.~(\ref{1.1.2.1}) for undirected graphs can be regarded as a specific case of Eq.~(\ref{directed_eq}) for directed graphs. The stationary distribution of the random walk in an undirected graph is $\pi(j) = D_{jj}/\sum_{i \in V} D_{ii}$. By substituting this expression into Eq.~(\ref{directed_eq}), we can easily derive Eq.~(\ref{1.1.2.1}), which is the exactly the regularizer of LGC by Zhou \etal~\cite{zhou2004learning}.

\subsection{Manifold regularization}
The manifold regularization~\cite{DBLP:journals/jmlr/BelkinNS06}~\cite{DBLP:journals/tnn/XuKLJ10} is actually a general framework that allows for developing a great number of algorithms ranging from supervised learning to unsupervised learning. However, it is viewed as a natural fit for GSSL since it combines the spectral graph theory with manifold learning to search for a low-dimensional representation with smoothness constraint in the original commonly high-dimensional space.

The manifold regularization framework fully utilizes the geometry property of the unknown probability distribution, which the data samples obey. Therefore, it introduces another term as a regularizer to control the complexity of the prediction function in the intrinsic space, measured by the geometry of the probability distribution. 

Formally, for a Mercer kernel $K\colon \mathcal{X} \times \mathcal{X} \rightarrow \mathbb{R}$, we denote the associated Reproducing Kernel Hilbert Space (RKHS) of the prediction function $f$. Then the loss function can be formulated in Eq.~(\ref{manifold_regularization}) as
\begin{equation}
    \label{manifold_regularization}
    \mathcal{L}(f)=\frac{1}{n_l}\sum_{i=1}^{n_l}\left(f(x_i) - y_i\right)^2 + \gamma_A \|f\|_{K}^{2} + \gamma_I \|f\|_{I}^{2},
\end{equation}
where $\gamma_A$ balances the complexity of the prediction function in the ambient space and $\gamma_I$ is the weighting parameter for the smoothness constraint term $\|f\|_{I}^{2}$ induced by both labeled and unlabeled samples.

It is noted that the added regularization term $\|f\|_{I}^{2}$ usually takes the following form,
\begin{equation}
\label{eq17}
    \|f\|_{I}^{2} = \frac{1}{(n_l + n_u)^2} \hat{y}^T L \hat{y},
\end{equation}
where $\hat{y} = \left[f(x_1),f(x_2),\dots,f(x_{n_l+n_u})\right]^T$ and $L$ is the Laplacian matrix of the graph.

According to the Representer Theorem~\cite{DBLP:journals/jmlr/ArgyriouMP09}, it is well-known that Eq.~(\ref{manifold_regularization}) has a closed-form solution when $\|f\|_{I}^{2}$ takes the form as shown in Eq.~(\ref{eq17}). However, it suffers from the high computational cost~\cite{DBLP:journals/jmlr/Niyogi13} which makes the algorithm unscalable when faced with large graphs. Popular solutions to alleviate this problem would be to accelerate either the construction of the Laplacian graph~\cite{DBLP:journals/jmlr/TalwalkarKMR13}~\cite{DBLP:conf/aaai/ZhangZYL16} or the kernel matrix operation~\cite{DBLP:journals/jmlr/ChangLZ17}~\cite{DBLP:conf/ijcai/LiLY019a}.

\subsection{LPDGL}
The above three methods~\cite{zhu2003semi}~\cite{zhou2004learning}~\cite{zhou2005icml} all prove to be ineffective for handling ambiguous examples~\cite{li2013adaptive}. Gong \etal~\cite{7010929} introduce deformed graph Laplacian (DGL) and provides the corresponding label prediction algorithm via DGL (LPDGL) for SSL. A new smoothness term that considers local information is added to the regularizer. The whole regularizer becomes Eq.~(\ref{LPDGL}) as
\begin{equation}
    \label{LPDGL}
   \begin{aligned}	
	\mathcal{L}(f)& =\frac{1}{2}\left(\alpha \sum_{i,j} W_{ij}\left(f(x_i) - f(x_j)\right)^{2}\right)\\ 
	& + \frac{1}{2}\left(\beta \sum_{i=1}^{n} (1 - \frac{D_{ii}}{\sum_{j=1}^{n}D_{jj}})\left(f(x_i)\right)^2\right) \\
	& + \mu \sum_{i=1}^{n_l}\left(f({x_i})-y_{i}\right)^{2},
\end{aligned}
\end{equation}
where both $\alpha$ and $\beta$ are trade-off parameters. It has been shown by theoretical analysis that LPDGL achieves a globally optimal prediction function. Additionally, the performance is robust to the hyper-parameters setting so this model is not difficult to fine-tune.

\subsection{Poisson learning}
The most recent work under the regularization framework is called poisson learning by~\cite{calder2020poisson}, which is motivated by the need to address the degeneracy of previous graph regularization methods when the label rate is meager. The new proposed approach replaces the given label values with the assignment of sources and sinks like flow in the graph. Thus, a resulting Poisson equation based on the graph can be nicely solved. The loss function of poisson learning is shown in Eq.~(\ref{poisson_learning}).
\begin{equation}
\label{poisson_learning}
   \begin{aligned}	
	\mathcal{L}(f)& =\frac{1}{2}\left(\sum_{i,j} W_{ij}\left(f(x_i) - \sum_{j \in \mathcal{N}(i)}f(x_j)\right)^{2}\right)\\ 
	& + \mu \sum_{i=1}^{n_l}\left(f({x_i})-y_{i}\right)^{2},
\end{aligned}
\end{equation}

\section{Graph embedding}
\label{sec:graph embedding}
Generally speaking, there are two types of graph embedding at two levels commonly seen in the literature. The first one is at the entire graph level while the second one is at the single node level~\cite{hamilton2017representation}. Both of them aim to represent the target object in a low-dimensional vector space. For GSSL, we focus on node embeddings since such representations can be easily used for SSL tasks. The main objective of node embedding is to encode the nodes as vectors with lower dimensions, which in the meantime can reflect their positions and the structure of their local neighborhood.

Formally, we have the following definition for node embedding on graphs. Given a graph $G=(V, E)$, a node embedding on it is a mapping $f_{\mathbf{z}}\colon v \rightarrow \mathbf{z}_v \in  \mathbb{R}^{d}$,  $\forall v \in {V}$ such that $d \ll|{V}|$ and the function $f_{\mathbf{z}}$ preserves some proximity measure defined on graph $G$. The generalized form of the loss function for graph embedding methods is shown in Eq.~(\ref{general_graph_embedding}) as
\begin{equation}
	\label{general_graph_embedding}
	\begin{aligned}
	\mathcal{L}(f) =& \sum_{(x_{i},y_{i}) \in \mathcal{D}_{l}}\mathcal{L}_{s}(f(f_{\mathbf{z}}(x_{i})),y_{i}) \\
	&+ \mu \sum_{x_i \in \mathcal{D}_l+\mathcal{D}_u}\mathcal{L}_{r}(f(f_{\mathbf{z}}(x_{i}))),
	\end{aligned}
\end{equation}
where $f_{\mathbf{z}}$ is the embedding function. It is obvious that Eq.~(\ref{general_graph_embedding}) is almost the same as Eq.~(\ref{general_graph_regularization}) for graph regularization except that for graph embedding methods, classifiers are trained based on the nodes' embedding results rather than nodes' attributes directly. 
\iffalse
An embedding therefore maps each node to a low-dimensional feature vector and tries to preserve the connection strengths between vertices. For instance, an embedding preserving first-order proximity might be obtained by minimizing $\sum_{i, j} s_{i j} \| y_{i}-y_{j}\|^{2}$. Let two node pairs $\left(v_{i}, v_{j}\right)$ and $\left(v_{i}, v_{k}\right)$ be associated with connections strengths such that $s_{i j}>s_{i k}$. In this case, $v_{i}$ and $v_{j}$ will be mapped to points in the embedding space that will be closer to each other than the mapping of $v_{i}$ and $v_{k}$.
\fi

\subsection{Generalization: Perspective of encoder-decoder}
\label{encode-decoder perspective}
Following the generalization methods on graph representation learning by Hamilton~\etal~\cite{hamilton2017representation}, all the node embedding methods mentioned in this section can be generalized under an encoder-decoder framework. From this perspective, the node embedding problem in graphs can be viewed as involving two key steps. First, an encoder model tries to map every node into a low-dimensional vector. Second, a decoder model is constructed to take the low-dimensional node embeddings as input and use them to reconstruct the information related to each node's neighborhood in the original graph, like an adjacency matrix.

\subsubsection{Encoder}
Formally, the encoder can be viewed as a function that maps nodes $v \in {V}$ to vector embeddings $\mathbf{z}_{v} \in \mathbb{R}^{d}$. The resulting embeddings are more discriminative in the latent space with more dimensions. Furthermore, they can be transformed back to the original feature vector more easily in the following decoder module. From a mathematical view, we have $\mathrm{Enc}\colon{V} \rightarrow \mathbb{R}^{d}.$

\subsubsection{Decoder}
The decoder module's main goal is to reconstruct certain graph statistics from the node embeddings generated by the encoder in the previous step. For example, given a node embedding $\mathbf{z}_{u}$ of a node $u$, the decoder might attempt to predict $u$'s set of neighbors $\mathcal{N}(u)$ or its row $\mathbf{A}[u]$ in the graph adjacency matrix.

Decoders are often defined in a pairwise form, which can be illustrated as predicting each pair of nodes' similarity. Formally, we have,
$\mathrm{Dec}\colon\mathbb{R}^{d} \times \mathbb{R}^{d} \rightarrow \mathbb{R}^{+}.$

\subsubsection{Reconstruction}
The reconstruction process of a pair of node embeddings $\mathbf{z}_{u},\mathbf{z}_{v}$ involves applying the pairwise decoder to them. The overall goal is to solve an optimization problem that minimizes the reconstruction loss so that the similarity measures produced by the decoder are as close to the ones defined in the original graph as possible. In a more formal way, we have
\begin{equation}
\label{reconstruction_goal}
    \operatorname{Dec}(\operatorname{Enc}(u), \operatorname{Enc}(v))=\operatorname{Dec}\left(\mathbf{z}_{u}, \mathbf{z}_{v}\right) \approx \mathbf{S}[u, v].
\end{equation}

Here, we assume that $\mathbf{S}[u, v]$ is a certain kind of similarity measure between a pair of nodes. For example, the commonly used reconstruction objective of predicting whether two nodes are neighbors would be minimizing the gap between $\mathbf{S}[u, v]$ and $\mathbf{A}[u, v]$.

To achieve the reconstruction objective Eq.~(\ref{reconstruction_goal}), the standard practice is to minimize the empirical reconstruction loss $\mathcal{L}$ defined for all the training data $\mathcal{D}$, including both labeled and unlabeled nodes,
\begin{equation}
\label{reconstruction_loss}
\mathcal{L}=\sum_{(u, v) \in \mathcal{D}} \ell\left(\operatorname{Dec}\left(\operatorname{Enc}[u], \operatorname{Enc}[v]\right), \mathbf{S}[u, v]\right),
\end{equation}
where $\ell\colon\mathbb{R} \times \mathbb{R} \rightarrow \mathbb{R}$ is a loss function for every node pair to compute the inconsistency between the true similarity values and the decoded ones.

\subsection{Shallow embedding and deep embedding}
In most of the work on node embedding, the encoder can be classified into a shallow embedding approach, in which this encoder is simply a lookup function based on the node ID. Additionally, the encoder can use both node features and the local graph structure around each node as the input to generate an embedding, like graph neural networks (GNNs). These methods are further categorized into the deep embedding method.

\section{Shallow Graph Embedding}
\label{shallow_embedding}
Some specialized optimization methods based on matrix factorization can be employed as a deterministic way to solve the optimization problem Eq.~(\ref{reconstruction_loss}). Generally speaking, the whole task can be considered as using matrix factorization methods to learn a low-dimensional approximation of a similarity matrix $\mathbf{S}$, where S encodes the information related to the original adjacency matrix or other matrix measurements.

Unlike the deterministic factorization methods, recent years have witnessed a surge in successful methods that use stochastic measures of neighborhood overlap to generate shallow embeddings. The key innovation in these approaches is that node embeddings are optimized under the assumption that if two nodes in the graph co-occur on some short-length random walks with high probability, they tend to share similar embeddings~\cite{spitzer2013principles}. 
% Figure \ref{outline_shallow_embedding} illustrates the outline of the reviewed approaches in Section~\ref{shallow_embedding}.
\iffalse
\begin{figure}[!t]
	\centering
	\includegraphics[width=0.35\textwidth]{"./pic/shallow_graph_embedding_methods_in_GSSL.pdf"}
	\caption{Shallow graph embedding methods in GSSL.}
	\label{outline_shallow_embedding}
\end{figure}
\fi
\subsection{Factorization-based methods}
For the category of factorization-based methods, a matrix that indicates the relationship between every node pair is factorized to obtain the node embedding. This matrix typically contains some underlying structural information about the constructed similarity graph, such as adjacency matrix and normalized Laplacian matrix. Different matrix properties can lead to different ways of factorizing these matrices. For instance, it is obvious that the normalized Laplacian matrix is positive semi-definite, so eigenvalue decomposition is a natural fit.

Table \ref{factorization_summary} applies the encoder-decoder perspective to summarize some representative factorization-based shallow embedding approach on node level for graphs. The most critical benefit of the previously mentioned encoder-decoder framework in Section~\ref{encode-decoder perspective} is that it provides a general overview of their respective components so that it is much easier to compare different embedding methods.

\begin{table*}[!t]
\caption{Summary on Factorization-based Shallow Graph Embedding Methods}
\centering
\begin{tabular}{ccccc}
\toprule
\textbf{Method}              & \textbf{Decoder}                                                 & \textbf{Similarity measure}                               & \textbf{Loss function}                                                                                           & \textbf{Time complexity}       \\           
\midrule
LLE~\cite{roweis2000nonlinear} & $\mathbf{z}_{u}-\sum_{v}{A}_{uv} \mathbf{z}_{v}$ & ${A}_{uv}$                                & $\sum_{u}\|\mathbf{z}_{u}-\sum_{v}{A}_{uv} \mathbf{z}_{v}\|^{2}$                                 & ${O}\left(|E| d^{2}\right)$ \\
Laplacian Eigenmaps~\cite{belkin2002laplacian} & $\left\|\mathbf{z}_{u}-\mathbf{z}_{v}\right\|_{2}^{2}$  & ${A}_{uv}$                              & $\operatorname{Dec}\left(\mathbf{z}_{u}, \mathbf{z}_{v}\right) \cdot \mathbf{S}[u, v]$                  & ${O}\left(|E| d^{2}\right)$ \\
Graph Factorization~\cite{ahmed2013distributed} & $\mathbf{z}_{u}^{\top} \mathbf{z}_{v}$                  & ${A}_{uv}$                               & $\left\|\operatorname{Dec}\left(\mathbf{z}_{u}, \mathbf{z}_{v}\right)-\mathbf{S}[u, v]\right\|_{2}^{2}$ & ${O}\left(|E| d\right)$     \\
GraRep~\cite{cao2015grarep}              & $\mathbf{z}_{u}^{\top} \mathbf{z}_{v}$                  & ${A}_{uv}, \ldots, {A}^{k}_{uv}$ & $\left\|\operatorname{Dec}\left(\mathbf{z}_{u}, \mathbf{z}_{v}\right)-\mathbf{S}[u, v]\right\|_{2}^{2}$ & ${O}\left(|V|^{3}\right)$   \\
HOPE~\cite{ou2016asymmetric}                & $\mathbf{z}_{u}^{\top} \mathbf{z}_{v}$                  & General Similarity Matrix $\mathbf{S}$  & $\left\|\operatorname{Dec}\left(\mathbf{z}_{u}, \mathbf{z}_{v}\right)-\mathbf{S}[u, v]\right\|_{2}^{2}$ & ${O}\left(|E| d^{2}\right)$ \\ \bottomrule
\label{factorization_summary}
\end{tabular}
\end{table*}

\subsubsection{Locally linear embedding (LLE)}
The most fundamental assumption in LLE~\cite{roweis2000nonlinear} is that the embedding result of each node is just a linear combination of the nodes in its neighborhood. More specifically, each entry $W_{ij}$ in the weight matrix $W$ for the constructed graph can denote how much the node $j$ contributes to the embedding of node $i$, namely the weight factor for the node $j$ in the linear combination of the node $i$. Formally, given the definition of $Y_{i}$, 
\begin{equation}Y_{i} \approx \sum_{j} W_{i j} Y_{j}, \quad \forall i \in V.\end{equation}
The embedding can be obtained as,
\begin{equation}\phi(Y)=\sum_{i}\left|Y_{i}-\sum_{j} W_{i j} Y_{j}\right|^{2}.\end{equation}

Adding another two constraints $\frac{1}{N} Y^{T} Y=I$ and $\sum_{i} Y_{i}=0$ into the above optimization equation, translational invariance can be eliminated since the embedding is forced to around the origin. It has been proven that the solution to this problem is to compute all the eigenvectors of the sparse matrix $(I-W)^{T}(I-W)$, sort the corresponding eigenvalues in the descending order and take the first $d + 1$ eigenvectors as the final embedding result.

\subsubsection{Laplacian eigenmaps}

Laplacian Eigenmaps~\cite{belkin2002laplacian} makes strongly connected nodes close to each other in the embedding space. Unlike the LLE~\cite{roweis2000nonlinear}, the objective function is designed in a pairwise manner,
\begin{equation}\begin{aligned}
\phi(Y) &=\frac{1}{2} \sum_{i, j}\left|Y_{i}-Y_{j}\right|^{2} W_{i j}, \\
&=\operatorname{tr}\left(Y^{T} L Y\right),
\end{aligned}\end{equation}
where $L$ is the Laplacian matrix. Similar to LLE~\cite{roweis2000nonlinear}, it is necessary to add another constraint $Y^{T} D Y=I$  so that some trivial solutions can be removed. The optimal solution is achieved by choosing the eigenvectors of the normalized Laplacian matrix whose corresponding eigenvalues are among the $d$ smallest ones.

\subsubsection{Graph factorization}
Graph Factorization (GF)~\cite{ahmed2013distributed} is the first algorithm to reduce the time complexity of previous graph embedding algorithms to $O(E)$. Instead of targeting at factorizing Laplacian matrix like LLE~\cite{roweis2000nonlinear} and Laplacian Eigenmaps~\cite{belkin2002laplacian}, GF directly employs the adjacency matrix and minimizes the objective function,
\begin{equation}\phi(Y, \mu)=\frac{1}{2} \sum_{(i, j) \in E}\left(W_{i j}-<Y_{i}, Y_{j}>\right)^{2}+\frac{\mu}{2} \sum_{i}\left\|Y_{i}\right\|^{2},\end{equation}
where $\mu$ is a hyper-parameter for the introduced regularization term. Because the adjacency matrix is not necessarily positive semidefinite, the summation over all the observed edges can be regarded as an approximation for the sake of scalability.

\subsubsection{GraRep}
GraRep~\cite{cao2015grarep} utilizes the node transition probability matrix which is defined as $T = D^{-1} W$ and $k$-order proximity is preserved by minimizing the loss $\|X^{k}-Y_{s}^{k} Y_{t}^{k T} \|_{F}^{2}$ where $X^{k}$ is derived from $T^{k}$, $Y_{s}$ and $Y_{t}$ are source and target embedding vectors respectively. It then concatenates $Y^{k}_{s}$ for all $k$ to form $Y_{s}$. The drawback of GraRep is scalability, since $T_{k}$ can have $O(|V|^{2})$ non-zero entries.

\subsubsection{HOPE}
Similar to GraRep~\cite{cao2015grarep}, HOPE~\cite{ou2016asymmetric} preserves higher-order proximity by minimizing another objective function $\left\|S-Y_{s} Y_{t}^{T}\right\|_{F}^{2}$, where $S$ is now the proximity matrix. The similarity measurement is defined in the form of $S=M_{g}^{-1} M_{l}$, where $M_{g}$ and $M_{l}$ are both sparse matrices. In this fashion, Singular Value Decomposition (SVD) can be applied so as to acquire node embeddings in an efficient manner.

\subsubsection{M-NMF}
While previous methods merely center around the microscopic structure (i.e., the first-order and second-order proximity), the mesoscopic community structure is incorporated into the embedding process for Modularized Nonnegative Matrix Factorization (M-NMF)~\cite{wang2017community}. The cooperation between the microscopic structure and the mesoscopic structure is established by exploiting the consensus relationship between the representations of nodes and the community structure.

\subsection{Random-walk-based methods}
The random walk is a powerful tool to gain approximate results about certain properties of the given graph, such as node centrality~\cite{newman2005measure} and similarity~\cite{fouss2007random}. Consequently, random-walk-based node embedding methods are effective under some scenarios when only part of the graph is accessible or the graph's scale is too large to handle efficiently.

The key points of random-walk-based node embeddings approaches are summarized in Table \ref{random_walk_summary_table} from the encoder-decoder perspective. The similarity function $p_{\mathcal{G}}(v \mid u)$ corresponds to the probability of visiting $v$ on a fixed-length random walk starting from $u$.

\begin{table*}[!ht]

\caption{Summary on Random-Walk-based Shallow Graph Embedding Methods}
\centering
\begin{tabular}{ccccc}
\toprule
\textbf{Method}   & \textbf{Decoder}                                                                                                              & \textbf{Similarity measure}          & \textbf{Loss function}                                                                                                                                                                                                                                          & \textbf{Time complexity}     \\ \midrule
DeepWalk~\cite{perozzi2014deepwalk} & $\frac{e^{\mathbf{z}_{u}^{\top} \mathbf{z}_{v}}}{\sum_{k \in {V}} e^{\mathbf{z}_{u}^{\top} \mathbf{z}_{k}}}$ & $p_{\mathcal{G}}(v \mid u)$ & $-\mathbf{S}[u, v] \log \left(\operatorname{Dec}\left(\mathbf{z}_{u}, \mathbf{z}_{v}\right)\right)$                                                                                                                                                    & ${O}(|V| d)$ \\
Planetoid~\cite{yang2016revisiting} & $\frac{e^{\mathbf{z}_{u}^{\top} \mathbf{z}_{v}}}{\sum_{k \in {V}} e^{\mathbf{z}_{u}^{\top} \mathbf{z}_{k}}}$ & $p_{\mathcal{G}}(v \mid u)$ & $\mathbb{E}_{v_{n} \sim P_{n}({V})}\left[\log \left(-\sigma\left(\mathbf{z}_{u}^{\top} \mathbf{z}_{v_{n}}\right)\right)\right]$ & ${O}(|V| d)$ \\ 
node2vec~\cite{perozzi2014deepwalk} & $\frac{e^{\mathbf{z}_{u}^{\top} \mathbf{z}_{v}}}{\sum_{k \in {V}} e^{\mathbf{z}_{u}^{\top} \mathbf{z}_{k}}}$ & $p_{\mathcal{G}}(v \mid u)$ & $\sum_{(u, v) \in \mathcal{D}}-\log \left(\sigma\left(\mathbf{z}_{u}^{\top} \mathbf{z}_{v}\right)\right)-\gamma \mathbb{E}_{v_{n} \sim P_{n}({V})}\left[\log \left(-\sigma\left(\mathbf{z}_{u}^{\top} \mathbf{z}_{v_{n}}\right)\right)\right]$ & ${O}(|V| d)$ \\ 
LINE~\cite{tang2015line} & $\frac{1}{1 - e^{-\mathbf{z}_{u}^{\top} \mathbf{z}_{k}}}$ & $p_{\mathcal{G}}(v \mid u)$ & $\sum_{(u, v) \in \mathcal{D}}-\log \left(\sigma\left(\mathbf{z}_{u}^{\top} \mathbf{z}_{v}\right)\right)-\gamma \mathbb{E}_{v_{n} \sim P_{n}({V})}\left[\log \left(-\sigma\left(\mathbf{z}_{u}^{\top} \mathbf{z}_{v_{n}}\right)\right)\right]$ & ${O}(|V| d)$ \\ 
PTE~\cite{tang2015pte} & $\frac{1}{1 - e^{-\mathbf{z}_{u}^{\top} \mathbf{z}_{k}}}$ & $p_{\mathcal{G}}(v \mid u)$ & $-\mathbf{S}[u, v] \log \left(p_{\mathcal{G}}(v \mid u)\right)$ & ${O}(|V| d)$ \\
\bottomrule
\label{random_walk_summary_table}
\end{tabular}
\end{table*}

\subsubsection{DeepWalk}
Inspired by the skip-gram model~\cite{mikolov2013distributed}, DeepWalk~\cite{perozzi2014deepwalk} follows the main goal of HOPE~\cite{ou2016asymmetric} and thus preserves higher-order proximity of each node pair. However, DeepWalk takes another approach by maximizing the possibility of encountering the previous $k$ nodes and the following $k$ nodes along one specific random walk with center $v_{i}$. In other words, DeepWalk maximizes the log-likelihood function which is defined as $\log \operatorname{Pr}\left(v_{i-k}, \ldots, v_{i-1}, v_{i+1}, \ldots, v_{i+k} \mid Y_{i}\right)$, where $2k + 1$ is the length of the random walk. The decoder is a basic form of a dot-product to reconstruct graph information from the encoded node embeddings.

\subsubsection{Planetoid}
Yang \etal~\cite{yang2016revisiting} propose another GSSL method based on the random walk, called Planetoid, where the embedding of a node is jointly trained to predict the class label and also the context in the given graph. The highlight of Planetoid is that it can be employed both in transductive and inductive settings. The context sampling behind Planetoid is actually built upon DeepWalk. In contrast to DeepWalk, Planetoid can handle graphs with real-value attributes by incorporating negative samples and injecting supervised information. The inductive variant of Planetoid views each node's embedding as a parameterized function of input feature vectors, while the transductive variant only embeds graph structure information.

\subsubsection{node2vec}
Following the same idea of DeepWalk~\cite{perozzi2014deepwalk}, node2vec~\cite{grover2016node2vec} also tries to preserve higher-order proximity for each node pair but makes full use of biased random walks so that it can balance between the breadth-first (BFS) and depth-first (DFS) search on the given graph to generate more expressive node embeddings. To be more specific, many random walks with fixed length are sampled, and then the possibility of occurrence of subsequent nodes along these biased random walks is maximized.

\subsubsection{LINE}
Previously mentioned methods do not scale in large real-world networks; Tang \etal propose LINE~\cite{tang2015line} to fix this issue by preserving both local and global graph structures with scalability. In particular, LINE combines first-order and second-order proximity, and they are optimized using the KL divergence metric. A decoder based on the sigmoid function is used in the first-order objective, while another decoder identical to the one in node2vec and DeepWalk is used in the second-order objective. Unlike node2vec and DeepWalk, LINE explicitly factorizes proximity measurement instead of implicitly incorporating it with sampled random walks.

\subsubsection{PTE}
PTE~\cite{tang2015pte} is proposed as a new semi-supervised representation learning method for text data. PTE fills the gap that many graph embedding methods are not particularly tuned for any task. The labeled information and various levels of information on word co-occurrence are first interpreted as a text network, which is then embedded into a low-dimensional space. This stochastic embedding method preserves the semantic meaning of words and shows a strong representational power for the particular downstream task. 

\subsubsection{HARP}
HARP~\cite{chen2017harp} is a general strategy to improve the above-mentioned solutions~\cite{perozzi2014deepwalk}~\cite{grover2016node2vec}~\cite{tang2015line} by avoiding local optima with the help of better weight initialization. The hierarchy of nodes is created by node aggregation in HARP using graph coarsening technique based on the preceding hierarchy layer. After that, the new embedding result can be generated from the coarsen graph, and the refined graph (i.e., the graph in the next level up in the hierarchy) can be initialized with the previous embedding. HARP propagates these node embeddings level by level so that it can be used in combination with random-walk-based approaches in order to achieve better performance.

\subsection{Relationship between random-walk-based and factorization-based methods}
Even though shallow embedding methods can be divided into two groups based on whether it is deterministic or stochastic, random-walk-based methods can actually be transformed into the factorization-based group in general. Qiu \etal~\cite{qiu2018network} provide a theoretical analysis of the aforementioned random-walk-based methods to show that they all essentially perform implicit matrix factorization and have closed-form solutions. Qiu \etal~\cite{qiu2018network} also propose a new framework, NetMF, to factorize these underlying matrices in random-walk-based methods explicitly. An impactful follow-up work, NetSMF~\cite{qiu-etal-2019-graph} extends NetMF~\cite{qiu2018network} to large-scale graphs based on sparse matrix factorization, making it more scalable for large networks in the real world.

\subsection{Limitations of shallow embedding}
\label{limitations_shallow_embedding}
Although shallow embedding methods have achieved impressive success on many SSL related tasks, it is worth noting that it also has some critical drawbacks that researchers found it hard to overcome with ease.
\begin{enumerate}
    \item \textbf{Lack of shared parameters.}
    In the encoder module, parameters are not shared between nodes since the encoder directly produces a unique embedding vector for each node. The lack of parameter sharing means that the number of parameters necessarily grows as $O(|{V}|)$, which can be intractable in massive graphs.
    \item \textbf{No use of node features.}
    Another key problem with shallow embedding approaches is that they fail to leverage node features. However, rich feature information could potentially be informative in the encoding process. This is especially true for SSL tasks where each node represents valuable feature information.
    \item \textbf{Failure in inductive applications.}
    Shallow embedding methods are inherently transductive~\cite{hamilton2017representation}. Generating embeddings for new nodes that are observed after the training phase is not possible. This restriction prevents shallow embedding methods from being used on inductive applications.
\end{enumerate}

\section{Deep Graph Embedding}
\label{sec:Deep Embedding}
In recent years, a great number of deep embedding approaches have been proposed to handle some of the limitations discussed in Section~\ref{limitations_shallow_embedding}. It should be emphasized that these deep embedding approaches differ from the shallow embedding approaches explained in Section~\ref{shallow_embedding} in that a much more complex encoder, which is often based on deep neural networks (DNN)~\cite{liu2017survey}, is constructed and employed. In this manner, the encoder module would incorporate both the structural and attribute information of the graph. For SSL tasks, a top-level classifier needs to be trained to predict class labels for unlabelled nodes under the transductive setting, based on the node embeddings generated by these deep learning models. 
\iffalse
Figure \ref{outline_deep_embedding} illustrates the outline of the reviewed approaches in Section~\ref{sec:Deep Embedding}.

\begin{figure}[!t]
	\centering
	\includegraphics[width=0.5\textwidth]{"./pic/deep_graph_embedding_methods_in_GSSL.pdf"}
	\caption{Deep graph embedding methods in GSSL.}
	\label{outline_deep_embedding}
\end{figure}
\fi
\subsection{AutoEncoder-based methods}
\begin{figure}[!t]
	\centering
	\includegraphics[width=0.45\textwidth]{"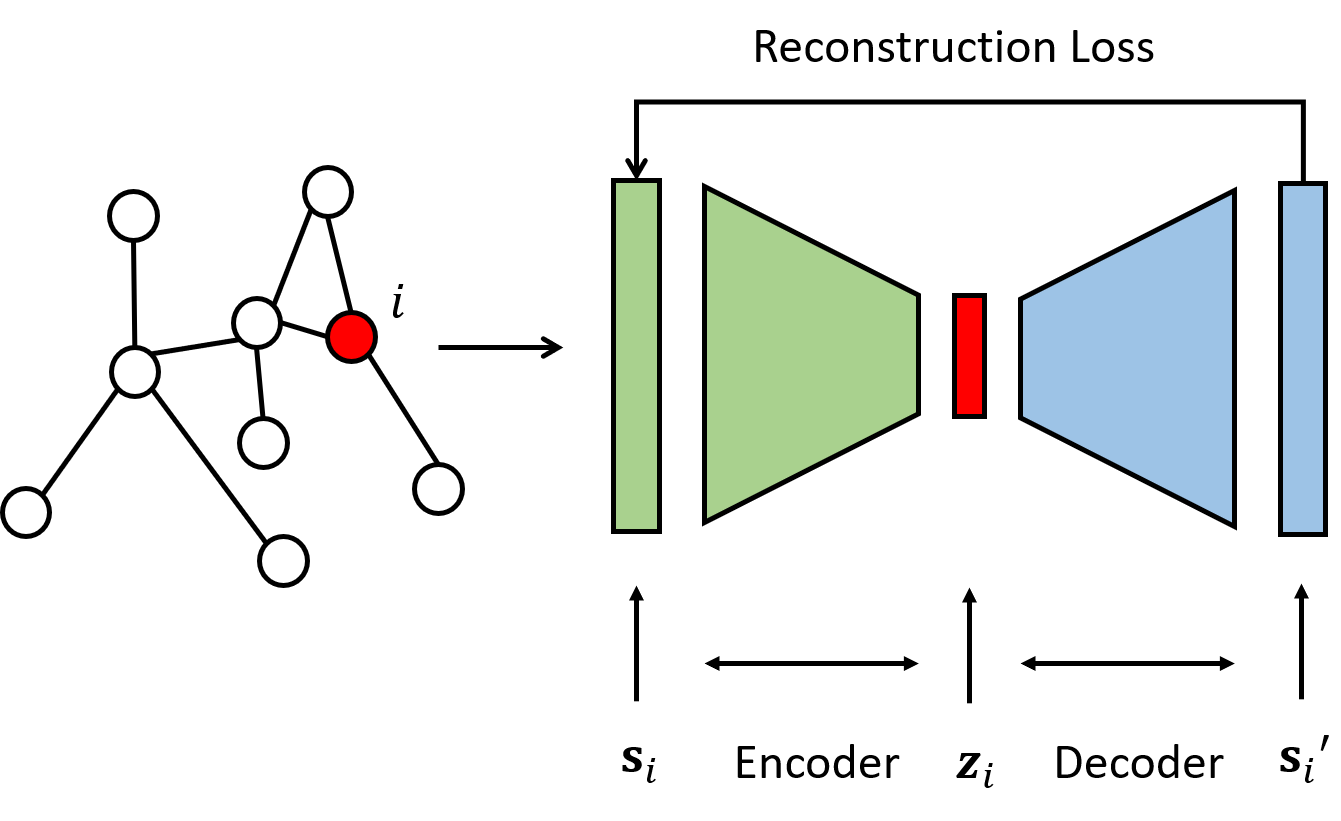"}
	\caption{For AutoEncoder-based methods, a high-dimensional vector $\mathbf{s}_i$ is extracted and fed into the AutoEncoder for generating a low-dimensional $\mathbf{z}_i$ embedding}
	\label{AE}
\end{figure}

Apart from the use of deep learning models, autoencoder-based methods also vary from the shallow embedding methods in that a unary decoder is employed instead of a pairwise one. Under the framework of autoencoder-based methods, every node, ${i}$, is represented by a high-dimensional vector extracted from a row in the similarity matrix, namely, $\mathbf{s}_{i} = {i}^{\text{th}} \text{ row of } \mathbf{S}$, where $\mathbf{S}_{i, j}=s_{\mathcal{G}}\left({i}, {j}\right)$. The autoencoder-based methods aims to first encode each node based on the corresponding vector $\mathbf{s}_{i}$ and then reconstruct it again from the embedding results, subject to the constraint that the reconstructed one should be as close to the original one as possible (Figure \ref{AE}):
\begin{equation}
\label{autoencoder_embedding_goal}
    \operatorname{Dec}\left(\operatorname{Enc}\left(\mathbf{s}_{i}\right)\right)=\operatorname{Dec}\left(\mathbf{z}_{i}\right) \approx \mathbf{s}_{i}.
\end{equation}
From the perspective of the loss function for autoencoder-based methods, it commonly keeps the following form:
\begin{equation}
\mathcal{L}=\sum_{{i} \in {V}}\left\|\operatorname{Dec}\left(\mathbf{z}_{i}\right)-\mathbf{s}_{i}\right\|_{2}^{2}.
\end{equation}

\begin{figure}[!t]
	\centering
	\includegraphics[width=0.45\textwidth]{"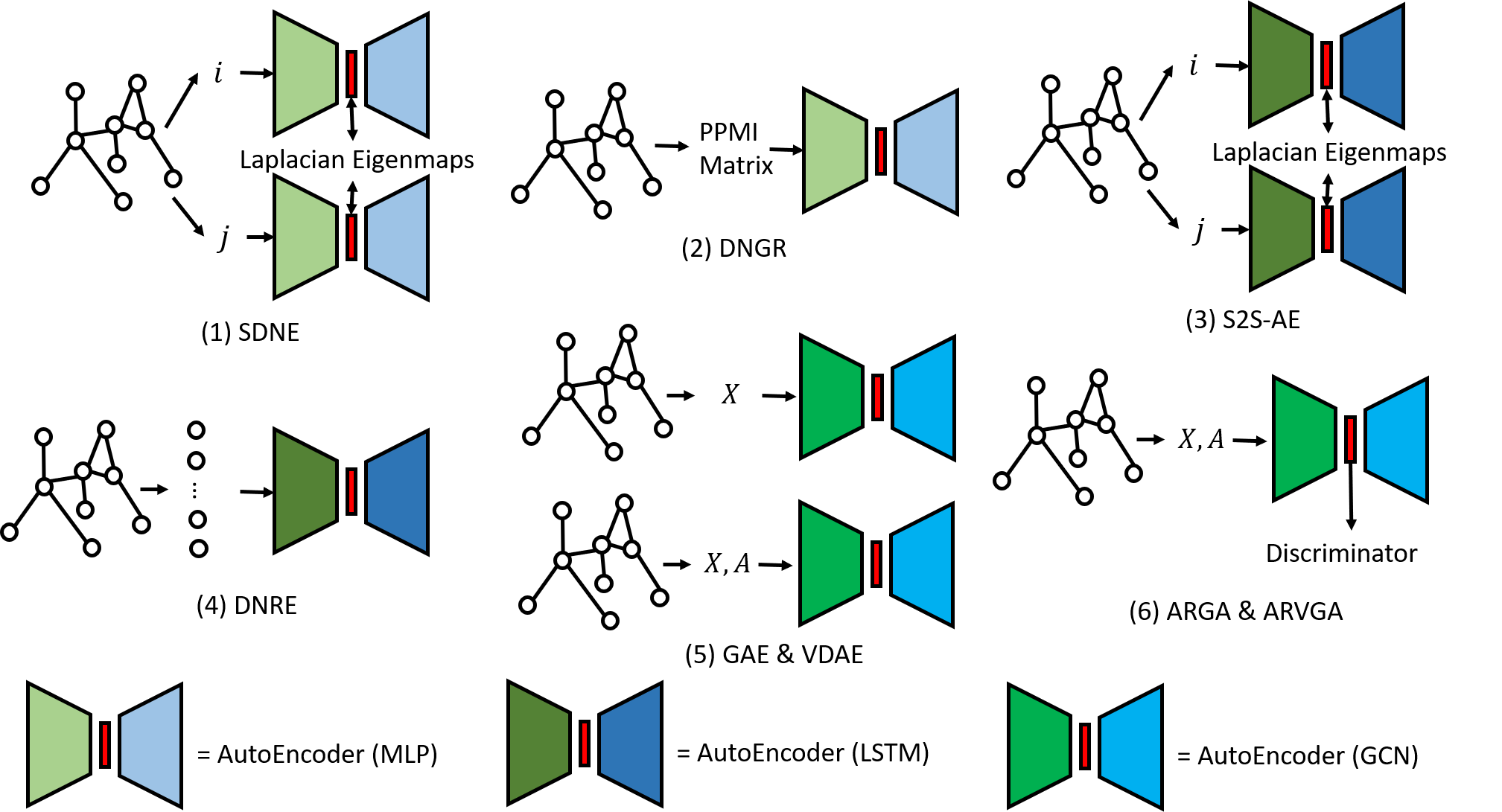"}
	\caption{Summary on the architectures of AutoEncoder-based methods}
	\label{AE_summary}
\end{figure}

From Eq.~(\ref{autoencoder_embedding_goal}), it should be pointed out that the encoder module actually depends on the given $ \mathbf{s}_{i}$ vector. This allows autoencoder-based deep embedding approaches to incorporate local structural information into the encoder, while it is simply impossible for the shallow embedding approaches to do so. The primary components of these methods are summarized as Table \ref{autoencoder_summary}, and the architectures of them are compared as Figure \ref{AE_summary}.

\begin{table*}[!t]
	\centering
	\caption{Summary on AutoEncoder-based Deep Graph Embedding Methods}
	\label{autoencoder_summary}
	\begin{tabular}{cccccc}
		\toprule
		\textbf{Method} &
		\textbf{Encoder} &
		\textbf{Deconder} &
		\textbf{Similarity measure} &
		\textbf{Loss function} &
		\textbf{Time complexity} \\
		\midrule
		SDNE~\cite{wang2016structural} &
		MLP &
		MLP &
		$\mathbf{s}_{u}$ &
		$\sum_{{u} \in {V}}\left\|\operatorname{Dec}\left(\mathbf{z}_{u}\right)-\mathbf{s}_{u}\right\|_{2}^{2}$ &
		${O}(|V \| E|)$ \\
		DNGR~\cite{cao2016deep} &
		MLP &
		MLP &
		$\mathbf{s}_{u}$ &
		$\sum_{{u} \in {V}}\left\|\operatorname{Dec}\left(\mathbf{z}_{u}\right)-\mathbf{s}_{u}\right\|_{2}^{2}$ &
		${O}\left(|V|^{2}\right)$ \\
		S2S-AE~\cite{taheri2018learning} &
		LSTM &
		LSTM &
		$\mathbf{s}_{u}$ &
		$\sum_{{u} \in {V}}\left\|\operatorname{Dec}\left(\mathbf{z}_{u}\right)-\mathbf{s}_{u}\right\|_{2}^{2}$ &
		${O}\left(|V|^{2}\right)$ \\
		DRNE~\cite{tu2018deep} &
		LSTM &
		LSTM &
		$\mathbf{s}_{u}$ &
		$\sum_{{u} \in {V}}\left\|\left(\mathbf{z}_{u}\right)-\sum_{v \in \mathcal{N}(u)} \operatorname{LSTM}\left(\mathbf{z}_{v}\right)\right\|_{2}^{2}$ &
		${O}(|V \| E|)$ \\
		GAE~\cite{kipf2016variational} &
		GCN &
		$\mathbf{z}_{u}^{\top} \mathbf{z}_{v}$ &
		$A_{uv}$ &
		$\sum_{{u} \in {V}}\left\|\operatorname{Dec}\left(\mathbf{z}_{u}\right)-{A}_{u}\right\|_{2}^{2}$ &
		${O}(|V \| E|)$ \\
		VGAE~\cite{kipf2016variational} &
		GCN &
		$\mathbf{z}_{u}^{\top} \mathbf{z}_{v}$ &
		$A_{uv}$ &
		$\mathbb{E}_{q(\mathbf{Z} \mid {X}, {A})}[\log p({A} \mid \mathbf{Z})]-\operatorname{KL}[q(\mathbf{Z} \mid {X}, {A}) \| p(\mathbf{Z})]$ &
		${O}(|V \| E|)$ \\
		ARGA~\cite{pan2019learning} &
		GAE &
		$\mathbf{z}_{u}^{\top} \mathbf{z}_{v}$ &
		$A_{uv}$ &
		$\min _{\mathcal{G}} \max _{\mathcal{D}} \mathbb{E}_{\mathbf{z} \sim p_{z}}[\log \mathcal{D}(\mathbf{Z})]+\mathbb{E}_{{x} \sim p({x})}[\log (1-\mathcal{D}(\mathcal{G}({X}, {A})))]$ &
		${O}(|V \| E|)$ \\
		ARVGA~\cite{pan2019learning} &
		VGAE &
		$\mathbf{z}_{u}^{\top} \mathbf{z}_{v}$ &
		$A_{uv}$ &
		$\min _{\mathcal{G}} \max _{\mathcal{D}} \mathbb{E}_{\mathbf{z} \sim p_{z}}[\log \mathcal{D}(\mathbf{Z})]+\mathbb{E}_{{x} \sim p({x})}[\log (1-\mathcal{D}(\mathcal{G}({X}, {A})))]$ &
		${O}(|V \| E|)$ \\
		\bottomrule
	\end{tabular}
\end{table*}

Despite this noticeable enhancement, the autoencoder-based methods may still suffer from some problems. Particularly, the computational cost of it is still intolerable for large scale graphs. Moreover, the structure of the autoencoder is predefined and unchanged during the training, so it is strictly transductive and thus fails to cope with evolving graphs. The up-to-date, relevant representative works to tackle these issues are~\cite{wang2017mgae}~\cite{ma2018constrained}.

\subsubsection{SDNE}
Wang \etal~\cite{wang2016structural} proposes Structural deep network integration (SDNE) with the help of deep autoencoders to preserve the proximity for first and second orders. The first-order proximity describes the similarity between each node pair, while the second-order proximity between each node pair describes the proximity of their neighborhood structure. The method takes advantage of non-linear functions to acquire the embedding results. It actually contains two modules: 1) the unsupervised part and 2) the supervised part. The former is an autoencoder designed to produce an embedding result for each node that can be used to rebuild its corresponding vector $\mathbf{s}_{i}$. For the latter part, Laplacian Eigenmaps is utilized so that penalty is imposed if connected nodes are encoded far away in the embedding space.

\subsubsection{DNGR}
Deep neural networks for learning graph representations (DNGR)~\cite{cao2016deep} integrates random surfing with autoencoders to generate node embeddings. This model has three components: 1) random surfing, 2) estimation of positive pointwise mutual information (PPMI) matrix and 3) stacked denoising autoencoders. For the input graph, random surfing is first applied to produce a co-occurrence probability matrix similar to HOPE. This probabilistic matrix is then converted into a PPMI matrix and fed into a stacked denoising autoencoder to generate the final embedding result. The feedback of the PPMI matrix guarantees the high order proximity is captured and maintained by the autoencoder. Moreover, the introduction of stacked denoising autoencoders enhances the model's robustness when the noise is present and the model's capability to detect the underlying structure required for some downstream tasks like node classification.

\subsubsection{S2S-AE}
Unlike previous methods whose encoders are all based on MLP, Taheri \etal~\cite{taheri2018learning} extend the form of the encoder to RNN models. S2S-AE~\cite{taheri2018learning} uses long short-term memory (LSTM)~\cite{hochreiter1997long} autoencoders to embed the graph sequences generated from random walks into a continuous vector space. The final representation is computed by averaging its graph sequence representations. The advantage of S2S-AE is that it can support arbitrary-length sequences, unlike others, which often suffer from the limitation of the fixed-length inputs.

\subsubsection{DRNE}
Deep recursive network embedding (DRNE)~\cite{tu2018deep} holds an assumption that the embedding of a node needs to approximate the aggregation of the embeddings of nodes within its neighborhood. It also uses LSTM~\cite{hochreiter1997long} to aggregate a node's neighbors, so the reconstruction loss is different from the one in S2S-AE~\cite{taheri2018learning} as suggested in Table \ref{autoencoder_summary}. In this way, DRNE can solve the issue that the LSTM model is not invariant when the given nodes' sequence permutes in different ways.

\subsubsection{GAE \& VGAE}
Both MLP-based and RNN-based methods only consider structural information and ignore the nodes' feature information. GAE~\cite{kipf2016variational} leverages GCN~\cite{kipf2016semi} to encode both. The encoder takes the form that,
\begin{equation}
    \label{vae_enc}
    \operatorname{Enc}({A},{X}) = \operatorname{GraphConv}\left(\sigma (\operatorname{GraphConv} ({A},{X}))\right),
\end{equation}
where $\operatorname{GraphConv}(\cdot)$ is a graph convolutional layer defined in~\cite{kipf2016semi}, $\sigma(\cdot)$ is the activation function, $A$ is the adjacency matrix, and $X$ is the attribute matrix. The decoder of GAE is defined as
\begin{equation}
    \label{vae_dec}
     \operatorname{Dec}{(\mathbf{z}_{u},\mathbf{z}_{v})} = \mathbf{z}_{u}^T \mathbf{z}_{v}.
\end{equation}
It may have some overfitting issue when the adjacency matrix is reconstructed in a direct way. Variational GAE (VGAE)~\cite{kipf2016variational} learns the distribution of data, in which the variational lower bound $\mathcal{L}$ is optimized.
\begin{equation}
    \label{vgae}
\mathcal{L}=\mathbb{E}_{q(\mathbf{Z} \mid {X}, {A})}[\log p({A} \mid \mathbf{Z})]-\operatorname{KL}[q(\mathbf{Z} \mid {X}, {A}) \| p(\mathbf{Z})],
\end{equation}
where $\operatorname{KL}[q(\cdot) \| p(\cdot)]$ is the Kullback-Leibler divergence between $q(\cdot)$ and $p(\cdot)$. Moreover, we have
\begin{equation}
q(\mathbf{Z} \mid {X}, {A})=\prod_{i=1}^{N} \mathcal{N}\left(\mathbf{z}_{i} \mid {\mu}_{i}, \operatorname{diag}\left({\sigma}_{i}^{2}\right)\right),
\end{equation}
and 
\begin{equation}
	p({A} \mid \mathbf{Z})=\prod_{i=1}^{N} A_{ij}\sigma\left(\mathbf{z}_{i}^{\top} \mathbf{z}_{j}\right) + \left(1-A_{ij}\right)\left(1-\sigma\left(\mathbf{z}_{i}^{\top} \mathbf{z}_{j}\right)\right).
\end{equation}
The most recent follow-up work are RWR-GAE~\cite{huang2019rwr} which adds a random walk regularizer for GAE and achieves noticeable performance improvement and DGVAE~\cite{DBLP:journals/corr/abs-2010-04408} which combines with graph cluster memberships as latent factors to further improve the internal mechanism of VAEs based graph generation.
\subsubsection{ARGA \& ARVGA}
To further improve the empirical distribution, $q(\mathbf{Z} \mid \mathbf{X}, \mathbf{A})$ in accordance with the prior distribution $p(\mathbf{A} \mid \mathbf{Z})$ in GAE and VGAE, Pan \etal~\cite{pan2019learning} propose ARGA and ARVGA with the help of the generative adversarial networks (GANs)~\cite{goodfellow2014generative}, in which they take GAE and VGAE as encoder respectively. 

\subsection{GNN-based methods}
\begin{figure}[!t]
	\centering
	\includegraphics[width=0.45\textwidth]{"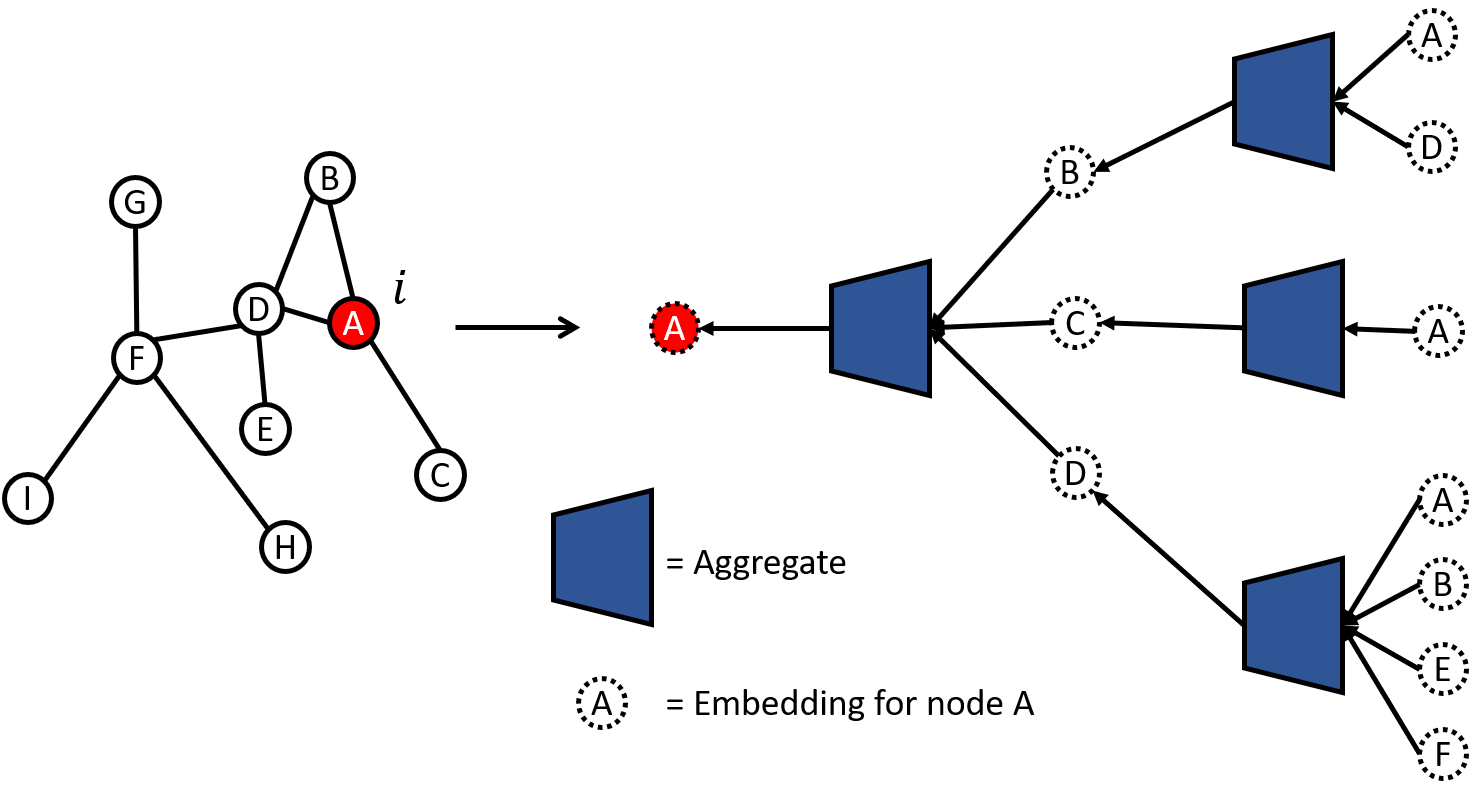"}
	\caption{GNN-based methods can generate node embeddings by aggregating embeddings from its neighbors}
	\label{GNN_illustration}
\end{figure}

% \textcolor{red}{more from GNN variants?}
Several up-to-date deep embedding approaches are designed to overcome the shallow embedding approaches' main drawbacks by constructing some specific functions that depend on a node's neighborhood (Figure \ref{GNN_illustration}). Graph neural network (GNN), which is heavily utilized in state-of-the-art deep embedding approaches, is considered as a general scheme for defining deep neural networks in the graph structure data. 

The main idea is that the representation vectors of nodes can depend not only on the structure of the graph but also on any feature information associated with the nodes. Dissimilar to the previously reviewed methods, graph neural networks use the node features, e.g., node information for a citation network or even simple statistics such as node degree, one-hot vectors, etc., to generate the desired node embeddings.

Like other deep node embedding methods, a classifier is trained on top of the node embeddings generated by the final hidden state in GNN-based models explicitly or implicitly. Afterward, it can be applied to the unlabeled nodes for SSL tasks. 

Since GNN consists of two main operations: Aggregate operation and Update operation, these methods will be reviewed from the perspective of the specific operation changed and improved compared with the basic GNN. The main techniques employed in these methods are also listed in Table \ref{GNN-based SSL} and some representative models.

\begin{table*}[!t]
    \caption{Summary on GNN-based Deep Graph Embedding Methods}
    \label{GNN-based SSL}
    \centering
\begin{tabular}{ccc}
\toprule
\textbf{Improvement} & \textbf{Technique}                      &\textbf{Model}                                   \\ \midrule
Basic GNN (Baseline)                             & Neural Message Passing         & Basic GNN~\cite{scarselli2008graph}          \\ \midrule
\multirow{4}{*}{Generalized Aggregate Operation} & Neighborhood Normalization     & GCN~\cite{kipf2016semi} MixHop~\cite{MixHop} SGC~\cite{SGC} DGN~\cite{DGN} \\
            & \multirow{2}{*}{Pooling}       & Set Pooling~\cite{wu2020comprehensive}   \\
            &                                & Janossy pooling~\cite{murphy2018janossy} \\
            & Neighborhood Attention         & GAT~\cite{velivckovic2017graph} AGNN~\cite{thekumparampil2018attention}         \\ \midrule
\multirow{5}{*}{Generalized Update Operation}    & \multirow{2}{*}{Concatenation} & Column networks~\cite{Pham2017AAAI} Scattering GCN~\cite{Scattering_GCN} \\
            &                                & GraphSAGE~\cite{hamilton2017inductive} DropEdge~\cite{DBLP:conf/iclr/RongHXH20}       \\
            & \multirow{2}{*}{Gated Updates} & GGNN~\cite{gatedGNN}                     \\
            &                                & NeuroSAT~\cite{lstmGNN}                  \\
            & JK connections                 & JK Networks~\cite{jkGNN}  InfoGraph*~\cite{DBLP:conf/iclr/SunHV020} \\ \bottomrule
\end{tabular}%
\end{table*}

\subsection{Basic GNN}
As Gilmer \etal~\cite{gilmer2017neural} point out, the fundamental feature of a basic GNN is that it takes advantages of \textit{neural message passing} in which messages are exchanged and updated between each pair of the nodes by using neural networks.

More specifically, during each neural message passing iteration in a basic GNN, a hidden embedding $\mathbf{h}_{u}^{(k)}$ corresponding to each node $u$ is updated according to message or information aggregated from $u$'s neighborhood $\mathcal{N}(u)$. This general message passing update rule can be expressed as follows:

% \begin{equation}
\begin{align}
\label{message_passing_update_rule}
&\mathbf{h}_{u}^{(k+1)} \nonumber \\
&=\texttt{Update}^{(k)}\left(\mathbf{h}_{u}^{(k)}, \texttt{Aggregate}^{(k)}\left(\left\{\mathbf{h}_{v}^{(k)}, \forall v \in \mathcal{N}(u)\right\}\right)\right), \nonumber \\
&=\texttt{Update}^{(k)}\left(\mathbf{h}_{u}^{(k)}, \mathbf{m}_{\mathcal{N}(u)}^{(k)}\right).
\end{align}
% \end{equation}
It is noteworthy that in Eq.~(\ref{message_passing_update_rule}), both the operation $\texttt{\textbf{Update}}$ and $\texttt{\textbf{Aggregate}}$ must be differentiable functions, typically, neural networks. Moreover, $\mathbf{m}_{\mathcal{N}(u)}$ is the exact message that is aggregated from node $u$'s neighborhood $\mathcal{N}(u)$ and tends to encode useful local structure information. Combining the message from neighbourhood with the previous hidden embedding state, the new state is generated according to Eq.~(\ref{message_passing_update_rule}). After a certain preset number of iterative steps, the last hidden embedding state converges so that this final state is regarded as the embedding output for each node. Formally, we have,
\begin{equation}\mathbf{z}_{u}=\mathbf{h}_{u}^{(K)}, \forall u \in \mathcal{V}.\end{equation}

It should be stressed that both the basic GNN and many of its variants strictly follow this generalized framework. The relationship among them is summarized as Table \ref{GNN-based SSL}.

Before the review on some of the GNN-based methods designed for SSL tasks, the basic version of GNN is introduced, which is a simplification of the original GNN model proposed by Scarselli \etal~\cite{scarselli2008graph}.

The basic GNN message passing is defined as:
\begin{equation}
\label{basic_gnn}\mathbf{h}_{u}^{(k)}=\sigma\left(\mathbf{W}_{\text {self }}^{(k)} \mathbf{h}_{u}^{(k-1)}+\mathbf{W}_{\text {neigh }}^{(k)} \sum_{v \in \mathcal{N}(u)} \mathbf{h}_{v}^{(k-1)}+\mathbf{b}^{(k)}\right),\end{equation}
where $\mathbf{W}_{\text {self }}^{(k)}, \mathbf{W}_{\text {neigh }}^{(k)}$ are trainable parameters and $\sigma$ is the activation function. 
The messages from the neighbors are firstly summarized. Then, the neighborhood information is combined together with the node's previous hidden embedding results by using a basic linear combination. Finally, a non-linearity activation function is applied on the combined information. From the perspective of the key components of the GNN framework, the Aggregation operation is and the Update operation is defined as shown in Eq.~(\ref{aggregate_gnn}) and Eq.~(\ref{update_gnn}).
\begin{equation}
\label{aggregate_gnn}\texttt{Aggregate}^{(k)}\left(\left\{\mathbf{h}_{v}^{(k)}, \forall v \in \mathcal{N}(u)\right\}\right) =\sum_{v \in \mathcal{N}(u)} \mathbf{h}_{v},\end{equation}

\begin{equation}
\label{update_gnn}\texttt{Update}\left(\mathbf{h}_{u}, \mathbf{m}_{\mathcal{N}(u)}\right)=\sigma\left(\mathbf{W}_{\text {self }} \mathbf{h}_{u}+\mathbf{W}_{\text {neigh }} \mathbf{m}_{\mathcal{N}(u)}\right).\end{equation}

Furthermore, it is not uncommon to add some self-loop tricks to the input graph so as to shut out the explicit update step, which can be considered as a straightforward simplification of the neural message passing method used in the basic GNN. To be a little more specific, the message passing process can now be simply defined as shown in Eq.~(\ref{self-loop-gnn}).
\begin{equation}
\label{self-loop-gnn}\mathbf{h}_{u}^{(k)}=\texttt{Aggregate}\left(\left\{\mathbf{h}_{v}^{(k-1)}, \forall v \in \mathcal{N}(u) \cup\{u\}\right\}\right).\end{equation}

As mentioned before, GNN models have all kinds of variants, which try to improve its performance and robustness to some extent. However, regardless of the variant of GNN, they all follow the neural message passing framework for Eq.~(\ref{message_passing_update_rule}) examined earlier. In the following two sections, Section~\ref{Generalized_Aggregation_Operation} and \ref{Generalized update operation}, some representative improvements on the two main components of basic GNN, aggregation operation and update operation, are reviewed in detail.

\subsection{Generalized aggregation operation}
\label{Generalized_Aggregation_Operation}
In general, the Aggregation operation in GNN models has received the most attention from the literature as a large number of researchers have proposed novel architectures or variations based on the original GNN model.

\subsubsection{Neighborhood normalization}
As previously stated, the most basic neighborhood aggregation operation, shown in Eq.~(\ref{aggregate_gnn}), solely computes the sum of the neighborhood's embedding states. The main problem with this approach is that it could be unstable and susceptible to the node's degree since nodes with a large degree tend to receive a large total value from more neighbors than those with fewer neighbors.

One typical and simple solution to this issue is to just normalize the aggregation operation based on the degree of the central nodes. The simplest approach is to just take an average rather than the sum by Eq.~(\ref{neighbor_norm_basic})
\begin{equation}
\label{neighbor_norm_basic}\mathbf{m}_{\mathcal{N}(u)}=\frac{\sum_{v \in \mathcal{N}(u)} \mathbf{h}_{v}}{|\mathcal{N}(u)|},\end{equation}
but methods with other normalization factors with similar ideas were proposed and achieved remarkable performance gain, such as the following symmetric normalization employed by Kipf \etal~\cite{kipf2016semi} in the GCN model as shown in Eq.~(\ref{gcn_neighbor_norm}).
\begin{equation}
\label{gcn_neighbor_norm}
\mathbf{m}_{\mathcal{N}(u)}=\sum_{v \in \mathcal{N}(u)} \frac{\mathbf{h}_{v}}{\sqrt{|\mathcal{N}(u)| \mid \mathcal{N}(v)\mid}}.\end{equation}

\textbf{Graph convolutional networks (GCNs)}.
One of the most popular and effective baseline GNN variants is the graph convolutional network (GCN)~\cite{kipf2016semi}, which is inspired by~\cite{bruna2013spectral} and~\cite{defferrard2016convolutional}. GCN makes full use of the neighborhood normalized aggregation techniques as well as the self-loop update operation. Therefore, the GCN model defines the update operation function as shown in Eq.~(\ref{GCN}). No aggregation operation is defined since it has been implicitly defined within the update operation function as
\begin{equation}
\label{GCN}
\mathbf{h}_{u}^{(k)}=\sigma\left(\mathbf{W}^{(k)} \sum_{v \in \mathcal{N}(u) \cup\{u\}} \frac{\mathbf{h}_{v}}{\sqrt{|\mathcal{N}(u)||\mathcal{N}(v)|}}\right).\end{equation}

There exist a great number of GCN variants to enhance SSL performance from different aspects. Li \etal~\cite{li2018deeper} are the first to provide deep insights into GCN's success and failure on SSL tasks. Later on, extensions to GCN for SSL begin to proliferate. Jiang \etal~\cite{jiang2019semi} explore the way to do graph construction based on GCN. Yang \etal~\cite{yang2020rethinking} combine the classic graph regularization methods with GCN. Abu \etal~\cite{abu2020n} present a novel N-GCN which marries the random walk with GCN, and a follow-up work GIL~\cite{DBLP:conf/iclr/XuCHZY020} with similar ideas is proposed as well. Other research work on GCN extensions can be found in~\cite{DBLP:conf/iclr/LiaoBTGUZ18}~\cite{zhang2019bayesian}~\cite{vashishth2019confidence}~\cite{10.1145/3178876.3186116}~\cite{wan2020contrastive}~\cite{kejani2020graph}~\cite{xu2020graph}~\cite{hu2019hierarchical}~\cite{DBLP:journals/corr/abs-1908-05081}.

\textbf{MixHop}.
GCN often fails to learn a generalized class of neighborhood with various mixing relationships. In order to overcome this limitation, MixHop~\cite{MixHop} is proposed to learn these relationships by repeatedly mixing feature representations of neighbors at various distances. Unlike GCN whose aggregation operator in the matrix form is defined as 
\begin{equation}
\label{GCN_matrix}
\mathbf{H}^{(k)}=\sigma\left(A\mathbf{H}^{(k-1)}\mathbf{W}^{(k)}\right),
\end{equation}
where $\mathbf{H}^{(k-1)}$ and $\mathbf{H}^{(k)}$ are the input and output hidden embedding matrix for layer $k$.
MixHop replaces the Graph Convolution (GC) layer
defined in Eq.~(\ref{GCN_matrix}) with
\begin{equation}
\label{MixHop_matrix}
\mathbf{H}^{(k)}=\|_{j \in P} \sigma\left({A}^{j} \mathbf{H}^{(k-1)} \mathbf{W}_{j}^{(i)}\right),
\end{equation}
where the hyper-parameter $P$ is a set of integer adjacency powers and $\|$ denotes column-wise concatenation. Specifically, by setting $P = \{1\}$, it exactly recovers the original GC layer. In fact, MixHop is interested in higher-order message passing, where each node receives latent representations from their immediate (one-hop) neighbors and from further N-hot neighbors.

\textbf{Simple graph convolution networks (SGC)}.
GCNs inherit unnecessary complexity and redundant computation cost in nature as it derives inspiration from deep learning methods. Wu \etal~\cite{SGC} reduce this excess complexity by eliminating the nonlinearities among every GCN layer and collapsing the original nonlinear function into a simple linear mapping function defined in Eq.~(\ref{SGC}). More importantly, these simplifications do not harm the prediction performance in many downstream applications.
\begin{equation}
\label{SGC}
\mathbf{h}_{u}^{(k)}=\sigma\left(\sum_{v \in \mathcal{N}(u) \cup\{u\}} \frac{\mathbf{h}_{v}}{\sqrt{|\mathcal{N}(u)||\mathcal{N}(v)|}}\right).
\end{equation}

\textbf{Differentiable group normalization (DGN)}.
To further mitigate the over-smoothing issue in GCN, DGN~\cite{DGN} also applies a new operation between the successive graph convolutional layers. Taking each embedding matrix $H^{(k)}$ generated from the $k^{th}$ graph convolutional layer as the input, DGN assigns each node into different groups and normalizes them independently to output a new embedding matrix for the next layer. Formally, we have,
\begin{equation}
    \label{DGN}
    \mathbf{H}^{(k+1)} = \mathbf{H}^{(k)} + \lambda \sum_{i=1}^{g}\left(\gamma_i(\frac{\mathbf{s}_{i}^{(k)}\circ\mathbf{H}^{(k)} - \mu_i}{\delta_i})+\beta_i\right),
\end{equation}
where $\mathbf{H}^{(k)}$ is the $k^{th}$ layer of the hidden embedding matrix, $\mathbf{s}_{i}$ is the similarity measure and $g$ is the total number of groups. In particular, $\mu_i$ and $\delta_i$ denote the vectors of running mean of group $i$, respectively, and $\gamma_i$ and $\beta_i$ denote the trainable scale and shift vectors, respectively.

\subsubsection{Pooling}
Aggregation operation is essentially a mapping from a set of neighborhood embedding results to a single vector with encoded information about the local structure and the feature of neighbor nodes' feature. In the previously reviewed settings of GNN models, the mapping function in the aggregation operation is simply the basic summation or linear functions over neighbor embeddings. Some more sophisticated and successful mapping functions used in the aggregation setting are reviewed in this section.

\textbf{Set pooling}.
In fact, according to Wu \etal~\cite{wu2020comprehensive}, one principal approach for designing an aggregation function is focused on the theory of permutation invariant neural networks. Generally, the permutation invariance property deals with problems concerning a set of objects: the target value for a given set is the same regardless of the order of the objects in the set. A typical example of an invariant permutation model is a convolutional neural network which performs the pooling operation over embedding extracted from a set's elements. Permutation invariance on graphs in general means that the aggregation function does not depend on the arbitrary order of the rows/columns in the adjacency matrix. For example, Zaheer \etal~\cite{zaheer2017deep} show that an aggregation function with the following form can be considered as a universal set function approximator:
\begin{equation}
\label{mlp_set_pooling}\mathbf{m}_{\mathcal{N}(u)}=\operatorname{MLP}_{\theta}\left(\sum_{v \in N(u)} \operatorname{MLP}_{\phi}\left(\mathbf{h}_{v}\right)\right).\end{equation}

\textbf{Janossy pooling}.
Another alternative method, called Janossy pooling, is to enhance the aggregation operation, which is also possibly more efficient than simply taking a sum or mean of the neighbor embeddings used in basic GNN. Janossy pooling~\cite{murphy2018janossy} uses a completely different approach. Instead of using a permutation invariant reduction (e.g., a sum or a mean), a permutation-sensitive function is applied, and the outcome is averaged over many potential permutations. 

Let $\pi_{i} \in \Pi$ denotes a permutation function that maps the set $\left\{\mathbf{h}_{v}, \forall v \in \mathcal{N}(u)\right\}$ to a specific sequence $\left(\mathbf{h}_{v_{1}}, \mathbf{h}_{v_{2}}, \ldots, \mathbf{h}_{v_{|\mathcal{N}(u)|}}\right)_{\pi_{i}}$. Namely, $\pi_{i} \in \Pi$ takes the unordered set of embedding states from the neighbors and puts them in a sequence dependent on some random ordering arbitrarily. The Janossy pooling approach then performs neighborhood aggregation operation by Eq.~(\ref{janossy_pooling}).
\begin{equation}
\label{janossy_pooling}
\mathbf{m}_{\mathcal{N}(u)}=\operatorname{MLP}_{\theta}\left(\frac{1}{|\Pi|} \sum_{\pi \in \Pi} \rho_{\phi}\left(\mathbf{h}_{v_{1}}, \mathbf{h}_{v_{2}}, \mathbf{h}_{v_{|\mathcal{N}(u)|}}\right)_{\pi_{i}}\right),\end{equation}
where $\Pi$ denotes a collection of permutations and $\rho_{\phi}$ is a permutation-sensitive function, e.g., a neural network that operates on sequential data. Usually $\rho_{\phi}$ is represented as an LSTM in operation, since LSTMs are known to be a powerful architecture for sequences in the neural network. 

\subsubsection{Neighborhood attention}
A common approach for enhancing the aggregation layer in GNNs is to implement some attention mechanisms~\cite{BahdanauCB14}, in addition to more general forms of set aggregation. The basic principle is to assign a weight or value of importance to each neighbor, which is used during the aggregation phase to weigh this neighbor's effect. 

\textbf{GAT}.
The first GNN model to apply this style of attention was Cucurull \etal's Graph Attention Network (GAT)~\cite{velivckovic2017graph}, which uses attention weights to define a weighted sum of the neighbors:
\begin{equation}\mathbf{m}_{\mathcal{N}(u)}=\sum_{v \in \mathcal{N}(u)} \alpha_{u, v} \mathbf{h}_{v},\end{equation}
where $\alpha_{u, v}$ denotes the attention on neighbor $v \in \mathcal{N}(u)$ when we are aggregating information at node $u$. In the original GAT paper, the attention weights are defined as
\begin{equation}\alpha_{u, v}=\frac{\exp \left(\mathbf{a}^{\top}\left[\mathbf{W} \mathbf{h}_{u} \oplus \mathbf{W} \mathbf{h}_{v}\right]\right)}{\sum_{v^{\prime} \in \mathcal{N}(u)} \exp \left(\mathbf{a}^{\top}\left[\mathbf{W h}_{u} \oplus \mathbf{W h}_{v^{\prime}}\right]\right)},\end{equation}
where $\mathbf{a}$ is a trainable attention vector, $\mathbf{W}$ is a trainable matrix, and denotes the concatenation operation. A similar parallel work is AGNN~\cite{thekumparampil2018attention} which reduces the number of parameters in GNN with the help of attention mechanisms.
\iffalse
Moreover, although it is less common in the GNN literature, the famous transformer~\cite{vaswani2017attention} may also have several ``heads`` for attention.
In this approach, one computes $K$ distinct attention weights $\alpha_{u, v, k}$ by using attention layers with independent parameters. The messages attached with the varying weights of attention are then transformed and combined with a linear projection, typically followed by a concatenation operation. Formally, we have,
\begin{equation}
    \mathbf{m}_{\mathcal{N}}(u) =\left[\mathbf{a}_{1} \oplus \mathbf{a}_{2} \oplus \ldots \oplus \mathbf{a}_{K}\right],
\end{equation}
\begin{equation}
    \mathbf{a}_{k} =\mathbf{W}_{i} \sum_{v \in \mathcal{N}(u)} \alpha_{u, v, k} \mathbf{h}_{v},
\end{equation}
where the attention weights $\alpha_{u, v, k}$ for each of the $K$ attention heads can be computed using some of the above mentioned attention mechanisms.

To sum up, a good strategic approach is incorporating the attention mechanism, particularly when some prior knowledge is available, as certain neighbors could be more knowledgeable than others, which may help increase the representational capacity of a GNN model.
\fi
\subsection{Generalized update operation}
\label{Generalized update operation}
As already noted in Section~\ref{Generalized_Aggregation_Operation}, lots of research papers focus on generalized aggregate operation. This was especially the case after the GraphSAGE Framework~\cite{hamilton2017inductive}, which implements the idea of generalized neighborhood aggregation. This section concentrates on the more diversified Update operation, which also makes the embeddings more suitable for SSL tasks.

\subsubsection{Concatenation and skip-connections}
Over-smoothing is a major issue for GNN. The over-smoothing is almost inevitable after many message iterations when the node-specific information becomes \enquote{washed away}. In such cases, the modified node representations are too highly dependent on the incoming message aggregated by the neighbors at the cost of previous layers' node hidden states. One reasonable way to mitigate this problem is to use vector concatenations or skip connections, which aim to retain information directly from previous rounds of the update.

These techniques can actually be used in combination with several other update operation methods for the GNN. For general purposes, $\texttt{Update}_{\texttt{base}}$ denotes the simple update rule that will be built on. For instance, the $\texttt{Update}_{\texttt{base}}$ function can be presumed as shown in Eq.~(\ref{update_gnn}) in the basic GNN.

\textbf{GraphSAGE}.
One of the simplest updates for skip connection is GraphSAGE~\cite{hamilton2017inductive} which uses a concatenation vector to hold more information from node level during message passing process:
\begin{equation}\texttt{Update}\left(\mathbf{h}_{u}, \mathbf{m}_{\mathcal{N}}(u)\right)=\left[\texttt {Update}_{\texttt{base}}\left(\mathbf{h}_{u}, \mathbf{m}_{\mathcal{N}(u)}\right) \oplus \mathbf{h}_{u}\right],\end{equation}
where the output from the simple update function is concatenated with the node 's previous layer representation. The core intuition is that the model is encouraged to dissociate information during the message passing.

\textbf{Column Network (CLN)}.
Besides concatenation methods, some other forms of skip-connections can also be applied, such as the linear interpolation method proposed by Pham \etal~\cite{Pham2017AAAI},
\begin{equation}
\begin{aligned}
\texttt{Update}\left(\mathbf{h}_{u}, \mathbf{m}_{\mathcal{N}}(u)\right)&=\boldsymbol{\alpha}_{1} \circ \texttt{Update}_{\texttt {base}}\left(\mathbf{h}_{u}, \mathbf{m}_{\mathcal{N}}(u)\right)\\
&+\boldsymbol{\alpha}_{2} \circ \mathbf{h}_{u},
\end{aligned}
\end{equation}
where $\boldsymbol{\alpha}_{1},\boldsymbol{\alpha}_{2} \in [0,1]^d$ are gating vectors with $\boldsymbol{\alpha}_{1}+\boldsymbol{\alpha}_{2}=1$ and $\circ$ denotes Hadamard product.
In this method the final update is a linear interpolation between the previous output and the current output and is modified depending on the information in the neighborhood.
\iffalse
In general, with the introduction of these concatenation and skip-connections techniques, not only the over-smoothing issue in GNN can be alleviated to some extent, but also the numerical stability of optimization is greatly improved. Indeed, similar to the functionality of residual connections in convolutional neural networks (CNNs), it is effective to apply these approaches to GNNs in which the training process of much deeper models could be facilitated.
\fi

\textbf{Scattering GCN}.
The most recent work on tackling the problem of over-smoothing in GNN is Scattering GCN~\cite{Scattering_GCN} with the geometric scattering transformation that enables band-pass filtering of graph signals. Geometric scattering is originally introduced in the context of whole-graph classification and consisted of aggregating scattering features. Similar and concurrent work is DropEdge~\cite{DBLP:conf/iclr/RongHXH20} which removes a certain portion of edges from the given graph at each training epoch, acting like a data augmenter and thus alleviate both over-smoothing and over-fitting issues at the same time.

These strategies are also beneficial for node classification tasks with relatively deep GNNs, in a semi-supervised setting, and they are excellent for these SSL tasks where the prediction in each node is closely correlated with the characteristics of the local neighborhood.

\subsubsection{Gated updates}
Parallel to the above-mentioned work, the researchers have also taken inspiration from the approaches used by recurrent neural networks (RNNs) to strengthen stability. One way to interpret the GNN message passing algorithm is to collect an observation from the neighbors from the aggregation operation, which then updates the hidden state of each node. From this perspective, some methods for updating the hidden status of RNN architectures can be directly applied based on the observation.

\textbf{GatedGNN}.
For example, one of the earliest GNN variants which put this idea into practice is proposed by Li \etal~\cite{gatedGNN}, in which the update operation is defined as shown in Eq.~(\ref{gatedGNN}) as,
\begin{equation}
\label{gatedGNN}
\mathbf{h}_{u}^{(k)}=\operatorname{GRU}\left(\mathbf{h}_{u}^{(k-1)}, \mathbf{m}_{\mathcal{N}(u)}^{(k)}\right),\end{equation}
where GRU is a gating mechanism function in recurrent neural networks, introduced by Kyunghyun Cho \etal~\cite{chung2014empirical}. Another approach called NeuroSAT~\cite{lstmGNN} has employed updates based on the LSTM architecture as well.
\iffalse
Typically every RNN update function can be used in the GNN setting.  Importantly, the RNN-style update parameters are also exchanged across nodes when updating each node based on the same LSTM or GRU cell. In reality, the model typically shares the parameters of the update operation function across the message passing layers.
\fi
\subsubsection{Jumping knowledge (JK) connections}
In the previous sections, it is implicitly assumed that the last layer's output is considered as the final embedding result. In other words, the node representations used for a downstream job, such as SSL tasks, are identical to the final layer's node embedding in the GNN. Formally, it is presumed that
\begin{equation}\mathbf{z}_{u}=\mathbf{h}_{u}^{(K)}, \forall u \in \mathcal{V}.\end{equation}

\textbf{JK Net}. 
A complementary approach to increase the effectiveness of final node representations is to use the combination on each layer of the message passing, rather than merely the final layer's output. In a more formal way,
\begin{equation}\mathbf{z}_{u}=f_{J K}\left(\mathbf{h}_{u}^{(0)} \oplus \mathbf{h}_{u}^{(1)} \oplus \ldots \oplus \mathbf{h}_{u}^{(K)}\right),\end{equation}
This technique is originally introduced and tested by Xu \etal~\cite{jkGNN}, called the idea of jumping knowledge connections. The $f_{JK}$ function can be used as the identity function for various applications, meaning that a simple concatenation is essentially performed among the node embeddings from each layer, but Xu \etal~\cite{jkGNN} also think about other possibilities such as max pooling. This method also leads to significant progress over a wide range of tasks like SSL classification and is usually regarded as an effective strategy.

\textbf{InfoGraph*}.
Previous GNN models appear to have low generalization performance because of the model's crafted styles. To lessen the downgrade of generalization performance in the testing phase, InfoGraph*~\cite{DBLP:conf/iclr/SunHV020} maximizes the mutual information between the embeddings learned by popular supervised learning methods and unsupervised representations learned by InfoGraph, where the given graph is encoded to produce its corresponding feature map by jumping knowledge concatenation. The encoder then learns from unlabeled samples while preserving the implicit semantic information that the downstream task favors.

\iffalse
\textbf{Summary.}
GNN-based methods are perceived as the most advanced approach and current trends for GSSL research. The most prominent advantage of GNN-based methods over others is that they consider nodes' attributes for generating embeddings so that the inconsistency in the proceeding graph construction step can be alleviated to some extent. Recent research work~\cite{verma2019graphmix} shows that GNN can be jointly trained with MLP based on a newly designed regularizer to boost SSL performance. Other related works focus on how to combine GNN with other models, like Markov model~\cite{DBLP:conf/icml/QuBT19}, GAN~\cite{DBLP:conf/cikm/DingTZ18} and metric learning~\cite{DBLP:conf/cvpr/LinGL20}. 

Finally, it is worth noting that the latest work on the GSSL method is GRAND~\cite{feng2020graphsemi}, which is considered as the state-of-the-art method by incorporating graph embedding with graph regularization. The key technique of GRAND is to introduce the classic data augmentation method on graph level into the GNN training as a pre-processing step so that the consistency regularization can be leveraged to optimize the prediction consistency of unlabeled data. GRAND is also able to mitigate the issues of over-smoothing and non-robustness further.
\fi
% \begin{adjustwidth}{-1in}{-1in}

% \end{adjustwidth}

\section{Applications}
\label{sec:applications}
\subsection{Datasets}
We summarize the commonly-used datasets in graph-based semi-supervised learning according to seven  different domains, namely citation networks, co-purchase networks, web page citation networks, and others. As shown in Table \ref{tab:summary_datasets} in Appendix \ref{dataset_sppendix}, the summary results on selected benchmark datasets on GSSL are listed with their respective statistical analysis.

\subsection{Open-source Implementations}
Here we list some open-source implementations for GSSL in Table \ref{tab:open_source_impl} in Appendix \ref{appendix_a}.

\subsection{Domains}
GSSL has a large number of successful applications across various domains. Some domains have graph-structure data in nature, while others do not. The former ones would be scenarios where raw data samples have explicit relational structure and can be easily constructed into a graph, such as traffic network in the cyber-physical systems (CPS), molecular structure in biomedical engineering, and friend recommendation in social networks. From the non-graph-structured data, however, a graph cannot be extracted directly. Typical examples would be more common scenarios, like image classification in computer vision and text classification in NLP.

\subsubsection{Computer vision}
Among many computer vision tasks (CV), hyperspectral image classification (HSI) is a representative example for GSSL applications. For one thing, labeled data in HSI is costly and scarce. For another, among all popular SSL methods, classic GSSL methods have elegant closed-form solutions and are easy to implement. Shao \etal~\cite{shao_probabilistic_2017}~\cite{shao_spatial_2018} propose a spatial and class structure regularized sparse representation graph for semi-supervised HSI classification. Later, Fang \etal~\cite{he_fast_2020} extend this work~\cite{shao_spatial_2018} by providing a more scalable algorithm based on anchor graph~\cite{chen2019deep}.

Pedronette \etal~\cite{pedronette_semi-supervised_2019} also improve the KNN-based graph construction methods~\cite{sslbook} in Section~\ref{sec:knn} to facilitate image retrieval. Another similar idea by Shi \etal~\cite{shi_graph_2020} is to use a temporal graph to assist image analysis.

The latest interesting work based on GSSL in CV is related to domain adaptation by He \etal~\cite{he_classification-aware_2020}. In this work, a novel idea of using graph-based manifold representation to do visual-audio transformation is proposed and examined.
\subsubsection{Natural language processing}
Amarnag \etal~\cite{10.5555/1870658.1870675} first introduce GSSL into traditional natural language processing (NLP) tasks and make pioneering work on part-of-speech (POS) tagging based on random fields~\cite{zhu2003semi}. The proposed algorithm uses a similarity graph to encourage similar n-grams to have similar POS tags. Later, Aliannejadi \etal~\cite{aliannejadi-etal-2014-graph} and Qiu \etal~\cite{qiu-etal-2019-graph} extend this work and use some GCN-based methods~\cite{kipf2016semi} to make the model more robust on other various natural language understanding (NLU) tasks.

More recent works on how to combine GSSL and NLP tasks center around graph smoothing problems. Mei \etal~\cite{mei2008general} propose a brand-new general optimization framework for smoothing language models with graph structures, which can further enhance the performance of information retrieval tasks. By constructing a similarity graph of documents and words, various types of GSSL methods can be performed.

Unlike the work~\cite{mei2008general} which studies the long texts, Hu \etal~\cite{linmei_heterogeneous_2019} focus on short texts in which the labeled data is sparse and limited. In particular, a flexible model based on GAT~\cite{velivckovic2017graph} with a dual-level attention mechanism is presented.

\subsubsection{Social networks}
A social network is a set of people with some pattern of interactions or \enquote{ties} between them and has graph-structured data explicitly. As is known to all, Twitter is one of the most famous and large-scale social networks, so various meaningful and interesting tasks based on GSSL can be performed on the Twitter dataset. Alam \etal~\cite{alam_graph_2018} adopt a graph-based deep learning framework by Yang \etal~\cite{pmlr-v48-yanga16} for learning an inductive semi-supervised model to classify tweets in a crisis situation. Later, Anand \etal~\cite{balaanand_enhanced_2019} improve classic GSSL methods to detect fake users from a large volume of Twitter data.

Another popular topic in social networks is related to POI recommendations, such as friend recommendation and follower suggestion. Yang \etal~\cite{10.1145/3097983.3098094} propose a general SSL framework to alleviate data scarcity via smoothing among users and POIs in the bipartite graph. Moreover, Chen \etal~\cite{chen2019deep} employ a user profiling approach to boost the performance of POI recommendation based heterogeneous graph attention networks by Wang \etal~\cite{HGAN}.

\subsubsection{Biomedical science}
Graphs are also ubiquitous in the area of biomedical science, such as the semantic biomedical knowledge graphs, molecular graphs for drugs, and protein-drug interaction for drug proposals. Doostparast \etal~\cite{doostparast_torshizi_graph-based_2018} use GSSL methods with genomic data integration to do phenotype classification tasks. In the meantime, Luo \etal~\cite{luo_semi-supervised_2018} provide a new graph regularization framework in heterogeneous networks to predict human miRNA-disease~\cite{zeng2019prediction}. Other typical applications are disease diagnosis~\cite{lang_graph-based_2020}, medical image segmentation~\cite{MAHAPATRA2017700} and medical Image classification~\cite{9095275}.

\section{Open problem}
\label{sec:open_problem}
Here a chronological overview of the mentioned representative methods in this survey is provided in Figure~\ref{fig:chronological_gssl}. From 2000 to 2012, the mainstream algorithm was centered around graph regularization and matrix factorization in the early years. After the resurgence of deep learning in 2015, the field witnessed the emergence of AutoEncoder-based methods while, in the meantime, random-walk-based methods coexisted. However, with the introduction of GCN~\cite{kipf2016semi} in 2017, the GNN-based method became the dominant solution and still is a heated topic now. Based on this trend, we list four potential research topics.

\subsection{Dynamicity}
Conventional GSSL methods reviewed in the preceding sections all treat the graph as a fixed observation. Ma \etal~\cite{ma2019flexible} is the first to apply generative models on GSSL. By viewing the graph as a random variable, the generated joint distribution can extract more general relationships among attributes, labels, and the graph structure. Moreover, this kind of model is more robust to missing data. Some latest follow-up works are \cite{pal2020non}~\cite{esmaeili2020new} and ~\cite{feng2020graph}. The most up-to-date work is \cite{DBLP:journals/corr/abs-2010-12783}, which provides a multi-source uncertainty framework for GNN and considers different types of uncertainties associated with class probabilities.

\subsection{Scalability}
Another open problem is how to make GSSL methods scalable when the input graph size increases rapidly. Pioneering work has been done by Liu \etal~\cite{liu2010large}~\cite{liu2012robust}, in which a novel graph is constructed with data points and anchor, namely, anchor graph regularization (AGR). Many successful follow-up works are proposed, such as~\cite{wang2016scalable}~\cite{he2020fast} for graph regularization methods and~\cite{verma2019stability}~\cite{li2020mixture} for GNN methods. These methods focus more on computational complexity instead of classification accuracy. It is worth noting that the most up-to-date work is GBP~\cite{DBLP:journals/corr/abs-2010-15421} which invents a new localized bidirectional propagation process from both the feature vectors and the nodes.

\subsection{Noise-resilience}
Graphs with noise or missing attributes are also heated topics. Most existing GSSL methods end up fully trusting the given few labels, but in real life, these labels are highly reliable as they are often produced by humans prone to mistakes. Stretcu \etal~\cite{stretcu2019graph} propose Graph Agreement Models (GAM), which introduces an auxiliary model that predicts the probability of two nodes sharing the same label. This co-training approach makes the model more resilient to noise. Zhao \etal~\cite{DBLP:journals/corr/abs-2010-12783} consider different types of uncertainties associated with class probabilities in the real case scenario. Similar latest works with the same goal but different approaches are~\cite{xu2020graph}~\cite{zhou2019graph}~\cite{de2020analysis}. In addition, Dunlop \etal~\cite{dunlop2020large} provide theoretical insights into this topic.

\subsection{Attack-robustness}
Robustness is always a common concern in machine learning systems. Liu \etal~\cite{liu2019unified} first propose a general framework for data poisoning attacks to GSSL. While ~\cite{liu2019unified} focuses more on how to generate a successful attack, \cite{liao2020graph} and \cite{DBLP:journals/corr/abs-2006-08149} defend GNN models from these adversarial attacks. Other relevant works via various approaches~\cite{gan2018safety}~\cite{DBLP:conf/icmcs/GaoHG20}~\cite{elinas2019variational} can be found as well.

\section{Conclusion}
To sum up, we conduct a comprehensive review of graph-based semi-supervised learning. A new taxonomy is proposed, in which all the popular label inference methods are grouped into two main categories: graph regularization and graph embedding. Moreover, they can be generalized within the regularization framework and the encoder-decoder framework respectively. Then a wide range of applications of GSSL are introduced along with relevant datasets, open-source codes for some of the GSSL methods. A chronological overview of these representative GSSL methods is presented in the Appendix. Finally, four open problems for future research directions are discussed as well.

\appendices

\section{Datasets collection for GSSL}
\label{dataset_sppendix}

\linespread{1.5}
\begin{table*}[!htp]
\centering
\caption{Summary of selected benchmark datasets on GSSL.}
\label{tab:summary_datasets}
{
\begin{tabular}{ccccccc}
\toprule
\textbf{Category}               & \textbf{Dataset}          & \textbf{\# Nodes}  & \textbf{\# Edges}   & \textbf{\# Features} & \textbf{\# Classes} & \textbf{Source}                             \\ \midrule
\multirow{4}{*}{Citation networks}         & Cora                  & 2,708  & 5,429   & 1,433 & 7   & \cite{sen2008collective}    \\
                       & Citeseer         & 3,327     & 4,732      & 3,703       & 6          & \cite{sen2008collective}           \\
                       & Pubmed           & 19,717    & 44,338     & 500         & 3          & \cite{sen2008collective}           \\
                       & DBLP             & 29,199    & 133,664    & 0           & 4          & \cite{perozzi2017don}          \\ \midrule
\multirow{4}{*}{Co-purchase networks}      & Amazon Computers      & 13,752 & 245,861 & 767   & 10  & \cite{shchur2018pitfalls}   \\
                       & Amazon Photo     & 7,650     & 119,081    & 745         & 8          & \cite{shchur2018pitfalls}          \\
                       & Coauthor CS      & 18,333    & 81,894     & 6,805       & 15         & \cite{shchur2018pitfalls}          \\
                       & Coauthor Physics & 34,493    & 247,962    & 8,415       & 5          & \cite{shchur2018pitfalls}          \\ \midrule
\multirow{4}{*}{Webpage citation networks} & Cornell               & 195    & 286     & 1,703 & 5   & \cite{wang2018dataset}      \\
                       & Texas            & 187       & 298        & 1,703       & 5          & \cite{wang2018dataset}             \\
                       & Washington       & 230       & 417        & 1,703       & 5          & \cite{wang2018dataset}             \\
                       & Wsicsonsin       & 265       & 479        & 1,703       & 5          & \cite{wang2018dataset}             \\ \midrule
\multirow{5}{*}{Social networks}           & Zachary’s karate club & 34     & 77      & 0     & 2   & \cite{girvan2002community}  \\
                       & Reddit           & 232965    & 11606919   & 602         & 41         & \cite{hamilton2017inductive}       \\
                       & BlogCatalog      & 10312     & 333983     & 0           & 39         & \cite{tang2009relational}          \\
                       & Flickr           & 1,715,256 & 22,613,981 & 0           & 5         & \cite{snapnets}          \\
                       & Youtube          & 1,138,499 & 2,990,443  & 0           & 47         & \cite{snapnets}          \\ \midrule
\multirow{1}{*}{Language networks}&Wikipedia &1,985,098 &1,000,924,086 & 0  & 7 & \cite{mahoney2011large} \\ \midrule                 
\multirow{6}{*}{Bio-chemical}              & PPI                   & 56944  & 818716  & 50    & 121 & \cite{zitnik2017predicting} \\
                       & NCI-1            & 29        & 32         & 37          & 2          & \cite{10.1007/s10115-007-0103-5}   \\
                       & MUTAG            & 17        & 19         & 7           & 2          & \cite{debnath1991structure}        \\
                       & D\&G             & 284       & 715        & 82          & 2          & \cite{dobson2003distinguishing}    \\
                       & PROTEIN          & 39        & 72         & 4           & 2          & \cite{borgwardt2005protein}        \\
                       & PTC              & 25        &            & 19          & 2          & \cite{toivonen2003statistical}     \\ \midrule
\multirow{1}{*}{Knowledge graph} & Nell & 65755     & 266144     & 61278       & 210        & \cite{carlson2010toward}           \\ \midrule
\multirow{3}{*}{Image} & MNIST            & 10,000    & 65,403     & 128         & 10         & \cite{deng2012mnist}               \\
                       & SVHN             & 10,000    & 68,844     & 128         & 10         & \cite{netzer2011reading}           \\
                       & CIFAR10          & 10,000    & 70,391     & 128         & 10         & \cite{krizhevsky2010convolutional} \\ \bottomrule
\end{tabular}
}
\end{table*}

% \clearpage
\section{Open-source Implementations}
\label{appendix_a}

\linespread{1.5}
\begin{table*}[!htp]
\centering
\caption{A Summary of Open-source Implementations}
\label{tab:open_source_impl}
{
\begin{tabular}{cccc}
\toprule
 &                        & \textbf{Model}                                           &\textbf{ Project Link }                                       \\ \midrule
\multirow{2}{*}{Graph Regularization} &
  \multirow{2}{*}{Label Propagation} &
  GRF~\cite{zhu2003semi} &
  \url{https://github.com/parthatalukdar/junto} \\
 &                        & LRC~\cite{zhou2004learning}                     & \url{https://github.com/provezano/lgc}              \\ \hline
\multirow{13}{*}{Shallow Graph Embedding} &
  \multirow{5}{*}{Matrix Factorization} &
  Laplacian Eigenmaps~\cite{belkin2002laplacian} &
  \url{https://github.com/thunlp/OpenNE} \\
 &                        & Graph Factorization~\cite{ahmed2013distributed} & \url{https://github.com/thunlp/OpenNE}              \\
 &                        & GraRep~\cite{cao2015grarep}                     & \url{https://github.com/benedekrozemberczki/GraRep} \\
 &                        & HOPE~\cite{ou2016asymmetric}                    & \url{https://github.com/ZW-ZHANG/HOPE}              \\
 &                        & M-NMF~\cite{wang2017community}                  & \url{https://github.com/benedekrozemberczki/M-NMF}  \\ \cline{2-4} 
 &
  \multirow{6}{*}{Random Walk} &
  DeepWalk~\cite{perozzi2014deepwalk} &
  \url{https://github.com/phanein/deepwalk} \\
 &                        & Planetoid~\cite{yang2016revisiting}             & \url{https://github.com/kimiyoung/planetoid}        \\
 &                        & node2vec~\cite{grover2016node2vec}              & \url{https://github.com/eliorc/node2vec}            \\
 &                        & LINE~\cite{tang2015line}                        & \url{https://github.com/tangjianpku/LINE}           \\
 &                        & PTE~\cite{tang2015pte}                          & \url{https://github.com/mnqu/PTE}                   \\
 &                        & HARP~\cite{chen2017harp}                        & \url{https://github.com/GTmac/HARP}                 \\ \cline{2-4} 
 & \multirow{2}{*}{Hybid} & NetMF~\cite{qiu2018network}                     & \url{https://github.com/xptree/NetMF}               \\
 &                        & NetSMF~\cite{qiu-etal-2019-graph}               & \url{https://github.com/xptree/NetSMF}              \\ \hline
\multirow{21}{*}{Deep Graph Embedding} &
  \multirow{6}{*}{AutoEncoder} &
  SDNE~\cite{wang2016structural} &
  \url{https://github.com/suanrong/SDNE} \\
 &                        & DNGR~\cite{cao2016deep}                         & \url{https://github.com/ShelsonCao/DNGR}            \\
 &                        & DRNE~\cite{tu2018deep}                          & \url{https://github.com/tadpole/DRNE}               \\
 &                        & GAE~\cite{kipf2016variational}                  & \url{https://github.com/tkipf/gae}                  \\
 &                        & VGAE~\cite{kipf2016variational}                 & \url{https://github.com/DaehanKim/vgae_pytorch}     \\
 &                        & ARGA~\cite{pan2019learning}                     & \url{https://github.com/Ruiqi-Hu/ARGA}              \\ \cline{2-4} 
 & \multirow{15}{*}{GNN}  & GCN~\cite{kipf2016semi}                         & \url{https://github.com/tkipf/gcn}                  \\
 &                        & MixHop~\cite{MixHop}                            & \url{https://github.com/samihaija/mixhop}           \\
 &                        & SGC~\cite{SGC}                                  & \url{https://github.com/Tiiiger/SGC}                \\
 &                        & DGN~\cite{DGN}                                  & \url{https://github.com/Kaixiong-Zhou/DGN}          \\
 &
   &
  Janossy pooling~\cite{murphy2018janossy} &
  \url{https://github.com/PurdueMINDS/JanossyPooling} \\
 &                        & GAT~\cite{velivckovic2017graph}                 & \url{https://github.com/PetarV-/GAT}                \\
 &                        & AGNN~\cite{thekumparampil2018attention}         & \url{https://github.com/dawnranger/pytorch-AGNN}    \\
 &                        & GraphSage~\cite{hamilton2017inductive}          & \url{https://github.com/williamleif/GraphSAGE}      \\
 &                        & DropEdge~\cite{DBLP:conf/iclr/RongHXH20}        & \url{https://github.com/DropEdge/DropEdge}          \\
 &                        & Column networks~\cite{Pham2017AAAI}             & \url{https://github.com/trangptm/Column_networks}   \\
 &                        & Scattering GCN~\cite{Scattering_GCN}            & \url{https://github.com/dms-net/scatteringGCN}      \\
 &                        & GGNN~\cite{gatedGNN}                            & \url{https://github.com/yujiali/ggnn}               \\
 &                        & NeuroSAT~\cite{lstmGNN}                         & \url{https://github.com/dselsam/neurosat}           \\
 &                        & JK Networks~\cite{jkGNN}         & \url{https://github.com/mori97/JKNet-dgl}           \\
 &                        & InfoGraph*~\cite{DBLP:conf/iclr/SunHV020}       & \url{https://github.com/fanyun-sun/InfoGraph}       \\ \bottomrule
\end{tabular}
}
\end{table*}

% \newpage
\section{Chronological overview of GSSL}
\begin{figure*}[!ht]
\centering
\scalebox{0.75}{
\begin{tikzpicture}[]
 
  % limits
  \newcount\yearOne; \yearOne=2000
  \def\w{20}      % width of axes
  \def\n{13}      % number of decades
  \def\noffset{1} % offset labels
  \def\nskip{3}   % skip number
  \def\la{2.00}   % arrow length
  \def\lt{0.20}   % tick length
  \def\ls{0.15}   % tick length (skipped)
 
  % help functions
  \def\myx(#1){{(#1-\yearOne)*\w/\n}}
    
  \def\arrowLabelRedDown(#1,#2,#3,#4){
    \def\yyp{{(\lt*(-0.10+#2))}}; \def\yyw{{(\yyp-\la*\lt*#3)}}
    \fill[red,radius=2pt] (\myx(#1),0) circle;
    \draw[<-,thick,red!50!red,align=center]
      (\myx(#1),\yyp) -- (\myx(#1),\yyw)
      node[below,red!80!red] {\strut#4}; %,fill=white
    }
    
\def\arrowLabelBlueUp(#1,#2,#3,#4){
    \def\yyp{{(\lt*(0.10+#2))}}; \def\yyw{{(\yyp+\la*\lt*#3)}}
    \fill[blue,radius=2pt] (\myx(#1),0) circle;
    \draw[<-,thick,blue!25!blue,align=center]
      (\myx(#1),\yyp) -- (\myx(#1),\yyw)
      node[above,blue!40!blue] {\strut#4}; %,fill=white
    }
    
    \def\arrowLabelCyanUp(#1,#2,#3,#4){
    \def\yyp{{(\lt*(0.10+#2))}}; \def\yyw{{(\yyp+\la*\lt*#3)}}
    \fill[cyan,radius=2pt] (\myx(#1),0) circle;
    \draw[<-,thick,cyan!25!cyan,align=center]
      (\myx(#1),\yyp) -- (\myx(#1),\yyw)
      node[above,cyan!40!cyan] {\strut#4}; %,fill=white
    }
    
      \def\arrowLabelForestGreenDown(#1,#2,#3,#4){
    \def\yyp{{(\lt*(-0.10+#2))}}; \def\yyw{{(\yyp-\la*\lt*#3)}}
    \fill[ForestGreen,radius=2pt] (\myx(#1),0) circle;
    \draw[<-,thick,ForestGreen!25!ForestGreen,align=center]
      (\myx(#1),\yyp) -- (\myx(#1),\yyw)
      node[below,ForestGreen!40!ForestGreen] {\strut#4}; %,fill=white
    }
 
  % axis
  \draw[->,thick] (-\w*0.03,0) -- (\w*1.06,0)
                  node[right=4pt,below=6pt] {\textbf{Year}};
 
  % ticks
  \foreach \tick in {0,1,...,\n}{
    \def\x{{\tick*\w/\n}}
    \def\year{\the\numexpr \yearOne+\tick*1 \relax}
% 	\pgfmathparse{Mod(\tick-\noffset,\nskip)==0?1:0}
% 	\ifnum\pgfmathresult>0
	  \draw[thick] (\x,\lt) -- (\x,-\lt) % ten tick
	               node[below] {\textbf{\year}}; % label
% 	\else
    %   \draw[thick] (\x,\ls) -- (\x,-\ls) % ten tick
                % node[below] {$10^{\dec}$}; % label
% 	\fi
  }

  % low mass
  \arrowLabelRedDown(2000.5,-1.2,3.0,LLE~\cite{roweis2000nonlinear})
\arrowLabelRedDown(2002.5,-1.2,3.0,Laplacian \\ Eigenmaps~\cite{belkin2002laplacian})
\arrowLabelForestGreenDown(2003.25,-1.2,2.0,GRF~\cite{zhu2003semi})
\arrowLabelForestGreenDown(2003.75,-1.2,3.0,LRC~\cite{zhou2004learning})
\arrowLabelForestGreenDown(2005.5,-1.2,3.1,Directed\\regularization~\cite{zhou2005icml})
\arrowLabelForestGreenDown(2006.5,-1.2,1.3,Manifold\\regularization~\cite{DBLP:journals/jmlr/BelkinNS06})
\arrowLabelBlueUp(2008.5,1.2,1.0,Basic\\GNN~\cite{scarselli2008graph})
\arrowLabelRedDown(2013.5,-1.2,3.0,GF~\cite{ahmed2013distributed})
%   \arrowLabelRed(-2.230,-1.2,3.0,$\text{B}'$) % ln(170) = 2.230
%   \arrowLabelBlue(-3.250,1.2,3.0,$\text{W}'$~\cite{9095275})
%   \arrowLabelBlue(-3.350,1.2,6.0,$\text{W}'$~\cite{9095275})
 
\end{tikzpicture}
}

\scalebox{0.75}{
\begin{tikzpicture}[]
 
  % limits
  \newcount\yearOne; \yearOne=2014
  \def\w{20}      % width of axes
  \def\n{3}      % number of decades
  \def\noffset{1} % offset labels
  \def\nskip{3}   % skip number
  \def\la{2.00}   % arrow length
  \def\lt{0.20}   % tick length
  \def\ls{0.15}   % tick length (skipped)
 
  % help functions
  \def\myx(#1){{(#1-\yearOne)*\w/\n}}
    
  \def\arrowLabelBurntOrangeDown(#1,#2,#3,#4){
    \def\yyp{{(\lt*(-0.10+#2))}}; \def\yyw{{(\yyp-\la*\lt*#3)}}
    \fill[BurntOrange,radius=2pt] (\myx(#1),0) circle;
    \draw[<-,thick,BurntOrange!50!BurntOrange,align=center]
      (\myx(#1),\yyp) -- (\myx(#1),\yyw)
      node[below,BurntOrange!80!BurntOrange] {\strut#4}; %,fill=white
    }
    
\def\arrowLabelRedDown(#1,#2,#3,#4){
    \def\yyp{{(\lt*(-0.10+#2))}}; \def\yyw{{(\yyp-\la*\lt*#3)}}
    \fill[Red,radius=2pt] (\myx(#1),0) circle;
    \draw[<-,thick,Red!50!Red,align=center]
      (\myx(#1),\yyp) -- (\myx(#1),\yyw)
      node[below,Red!80!Red] {\strut#4}; %,fill=white
    }

  \def\arrowLabelBlueUp(#1,#2,#3,#4){
    \def\yyp{{(\lt*(0.10+#2))}}; \def\yyw{{(\yyp+\la*\lt*#3)}}
    \fill[blue,radius=2pt] (\myx(#1),0) circle;
    \draw[<-,thick,blue!25!blue,align=center]
      (\myx(#1),\yyp) -- (\myx(#1),\yyw)
      node[above,blue!40!blue] {\strut#4}; %,fill=white
    }
    
    \def\arrowLabelCyanUp(#1,#2,#3,#4){
    \def\yyp{{(\lt*(0.10+#2))}}; \def\yyw{{(\yyp+\la*\lt*#3)}}
    \fill[cyan,radius=2pt] (\myx(#1),0) circle;
    \draw[<-,thick,cyan!25!cyan,align=center]
      (\myx(#1),\yyp) -- (\myx(#1),\yyw)
      node[above,cyan!40!cyan] {\strut#4}; %,fill=white
    }
    
      \def\arrowLabelForestGreenDown(#1,#2,#3,#4){
    \def\yyp{{(\lt*(-0.10+#2))}}; \def\yyw{{(\yyp-\la*\lt*#3)}}
    \fill[ForestGreen,radius=2pt] (\myx(#1),0) circle;
    \draw[<-,thick,ForestGreen!25!ForestGreen,align=center]
      (\myx(#1),\yyp) -- (\myx(#1),\yyw)
      node[below,ForestGreen!40!ForestGreen] {\strut#4}; %,fill=white
    }
 
  % axis
  \draw[->,thick] (-\w*0.03,0) -- (\w*1.06,0)
                  node[right=4pt,below=6pt] {\textbf{Year}};
 
  % ticks
  \foreach \tick in {0,1,...,\n}{
    \def\x{{\tick*\w/\n}}
    \def\year{\the\numexpr \yearOne+\tick*1 \relax}
% 	\pgfmathparse{Mod(\tick-\noffset,\nskip)==0?1:0}
% 	\ifnum\pgfmathresult>0
	  \draw[thick] (\x,\lt) -- (\x,-\lt) % ten tick
	               node[below] {\textbf{\year}}; % label
% 	\else
    %   \draw[thick] (\x,\ls) -- (\x,-\ls) % ten tick
                % node[below] {$10^{\dec}$}; % label
% 	\fi
  }

  % low mass
\arrowLabelBurntOrangeDown(2014.3,-1.2,3.0,DeepWalk~\cite{perozzi2014deepwalk})
\arrowLabelForestGreenDown(2015.2,-1.2,3.0,LPDGL~\cite{7010929})
\arrowLabelRedDown(2015.4,-1.2,1.5,GraRep~\cite{cao2015grarep})
\arrowLabelBurntOrangeDown(2015.6,-1.2,3.0,LINE~\cite{tang2015line})
\arrowLabelBurntOrangeDown(2015.9,-1.2,2.0,PTE~\cite{tang2015pte})
\arrowLabelRedDown(2016.1,-1.2,3.0,HOPE~\cite{ou2016asymmetric})
\arrowLabelBurntOrangeDown(2016.3,-1.2,2.0,Planetoid~\cite{yang2016revisiting})
\arrowLabelBurntOrangeDown(2016.7,-1.2,3.0,node2vec~\cite{perozzi2014deepwalk})
\arrowLabelCyanUp(2016.2,1.2,2.0,SDNE~\cite{wang2016structural})
\arrowLabelCyanUp(2016.4,1.2,1.0,DNGR~\cite{cao2016deep})
\arrowLabelCyanUp(2016.6,1.2,2.0,GAE~\cite{kipf2016variational}\\GVAE~\cite{kipf2016variational})
\arrowLabelBlueUp(2016.8,1.2,1.0,Gated GNN~\cite{gatedGNN})
\end{tikzpicture}
}

\scalebox{0.75}{
\begin{tikzpicture}[]
 
  % limits
  \newcount\yearOne; \yearOne=2017
  \def\w{20}      % width of axes
  \def\n{2}      % number of decades
  \def\noffset{1} % offset labels
  \def\nskip{3}   % skip number
  \def\la{2.00}   % arrow length
  \def\lt{0.20}   % tick length
  \def\ls{0.15}   % tick length (skipped)
 
  % help functions
  \def\myx(#1){{(#1-\yearOne)*\w/\n}}
    
  \def\arrowLabelBurntOrangeDown(#1,#2,#3,#4){
    \def\yyp{{(\lt*(-0.10+#2))}}; \def\yyw{{(\yyp-\la*\lt*#3)}}
    \fill[BurntOrange,radius=2pt] (\myx(#1),0) circle;
    \draw[<-,thick,BurntOrange!50!BurntOrange,align=center]
      (\myx(#1),\yyp) -- (\myx(#1),\yyw)
      node[below,BurntOrange!80!BurntOrange] {\strut#4}; %,fill=white
    }
    
\def\arrowLabelRedDown(#1,#2,#3,#4){
    \def\yyp{{(\lt*(-0.10+#2))}}; \def\yyw{{(\yyp-\la*\lt*#3)}}
    \fill[Red,radius=2pt] (\myx(#1),0) circle;
    \draw[<-,thick,Red!50!Red,align=center]
      (\myx(#1),\yyp) -- (\myx(#1),\yyw)
      node[below,Red!80!Red] {\strut#4}; %,fill=white
    }

  \def\arrowLabelBlueUp(#1,#2,#3,#4){
    \def\yyp{{(\lt*(0.10+#2))}}; \def\yyw{{(\yyp+\la*\lt*#3)}}
    \fill[blue,radius=2pt] (\myx(#1),0) circle;
    \draw[<-,thick,blue!25!blue,align=center]
      (\myx(#1),\yyp) -- (\myx(#1),\yyw)
      node[above,blue!40!blue] {\strut#4}; %,fill=white
    }
    
    \def\arrowLabelCyanUp(#1,#2,#3,#4){
    \def\yyp{{(\lt*(0.10+#2))}}; \def\yyw{{(\yyp+\la*\lt*#3)}}
    \fill[cyan,radius=2pt] (\myx(#1),0) circle;
    \draw[<-,thick,cyan!25!cyan,align=center]
      (\myx(#1),\yyp) -- (\myx(#1),\yyw)
      node[above,cyan!40!cyan] {\strut#4}; %,fill=white
    }
    
      \def\arrowLabelForestGreenDown(#1,#2,#3,#4){
    \def\yyp{{(\lt*(-0.10+#2))}}; \def\yyw{{(\yyp-\la*\lt*#3)}}
    \fill[ForestGreen,radius=2pt] (\myx(#1),0) circle;
    \draw[<-,thick,ForestGreen!25!ForestGreen,align=center]
      (\myx(#1),\yyp) -- (\myx(#1),\yyw)
      node[below,ForestGreen!40!ForestGreen] {\strut#4}; %,fill=white
    }
 
  % axis
  \draw[->,thick] (-\w*0.03,0) -- (\w*1.06,0)
                  node[right=4pt,below=6pt] {\textbf{Year}};
 
  % ticks
  \foreach \tick in {0,1,...,\n}{
    \def\x{{\tick*\w/\n}}
    \def\year{\the\numexpr \yearOne+\tick*1 \relax}
% 	\pgfmathparse{Mod(\tick-\noffset,\nskip)==0?1:0}
% 	\ifnum\pgfmathresult>0
	  \draw[thick] (\x,\lt) -- (\x,-\lt) % ten tick
	               node[below] {\textbf{\year}}; % label
% 	\else
    %   \draw[thick] (\x,\ls) -- (\x,-\ls) % ten tick
                % node[below] {$10^{\dec}$}; % label
% 	\fi
  }

  % low mass
\arrowLabelRedDown(2017.2,-1.2,2.0,M-NMF~\cite{wang2017community})
\arrowLabelBlueUp(2017.3,1.2,2.0,GCN~\cite{kipf2016semi})
\arrowLabelBlueUp(2017.5,1.2,1.0,Deep Set~\cite{zaheer2017deep})
\arrowLabelBlueUp(2017.7,1.2,2.0,GraphSAGE~\cite{hamilton2017inductive})
\arrowLabelBlueUp(2017.9,1.2,2.0,CLN~\cite{Pham2017AAAI})
\arrowLabelBurntOrangeDown(2018.1,-1.2,2.0,HARP~\cite{chen2017harp})
\arrowLabelCyanUp(2018.2,1.2,2.0,S2S-AE~\cite{taheri2018learning})
\arrowLabelCyanUp(2018.4,1.2,2.0,DRNE~\cite{tu2018deep})
\arrowLabelBlueUp(2018.6,1.2,2.0,GAT~\cite{velivckovic2017graph})
\arrowLabelBlueUp(2018.8,1.2,2.0,JK Networks~\cite{jkGNN})
\end{tikzpicture}
}

\scalebox{0.75}{
\begin{tikzpicture}[]
 
  % limits
  \newcount\yearOne; \yearOne=2019
  \def\w{20}      % width of axes
  \def\n{2}      % number of decades
  \def\noffset{1} % offset labels
  \def\nskip{3}   % skip number
  \def\la{2.00}   % arrow length
  \def\lt{0.20}   % tick length
  \def\ls{0.15}   % tick length (skipped)
 
  % help functions
  \def\myx(#1){{(#1-\yearOne)*\w/\n}}
    
  \def\arrowLabelBurntOrangeDown(#1,#2,#3,#4){
    \def\yyp{{(\lt*(-0.10+#2))}}; \def\yyw{{(\yyp-\la*\lt*#3)}}
    \fill[BurntOrange,radius=2pt] (\myx(#1),0) circle;
    \draw[<-,thick,BurntOrange!50!BurntOrange,align=center]
      (\myx(#1),\yyp) -- (\myx(#1),\yyw)
      node[below,BurntOrange!80!BurntOrange] {\strut#4}; %,fill=white
    }
    
\def\arrowLabelRedDown(#1,#2,#3,#4){
    \def\yyp{{(\lt*(-0.10+#2))}}; \def\yyw{{(\yyp-\la*\lt*#3)}}
    \fill[Red,radius=2pt] (\myx(#1),0) circle;
    \draw[<-,thick,Red!50!Red,align=center]
      (\myx(#1),\yyp) -- (\myx(#1),\yyw)
      node[below,Red!80!Red] {\strut#4}; %,fill=white
    }

  \def\arrowLabelBlueUp(#1,#2,#3,#4){
    \def\yyp{{(\lt*(0.10+#2))}}; \def\yyw{{(\yyp+\la*\lt*#3)}}
    \fill[blue,radius=2pt] (\myx(#1),0) circle;
    \draw[<-,thick,blue!25!blue,align=center]
      (\myx(#1),\yyp) -- (\myx(#1),\yyw)
      node[above,blue!40!blue] {\strut#4}; %,fill=white
    }
    
    \def\arrowLabelCyanUp(#1,#2,#3,#4){
    \def\yyp{{(\lt*(0.10+#2))}}; \def\yyw{{(\yyp+\la*\lt*#3)}}
    \fill[cyan,radius=2pt] (\myx(#1),0) circle;
    \draw[<-,thick,cyan!25!cyan,align=center]
      (\myx(#1),\yyp) -- (\myx(#1),\yyw)
      node[above,cyan!40!cyan] {\strut#4}; %,fill=white
    }
    
      \def\arrowLabelForestGreenDown(#1,#2,#3,#4){
    \def\yyp{{(\lt*(-0.10+#2))}}; \def\yyw{{(\yyp-\la*\lt*#3)}}
    \fill[ForestGreen,radius=2pt] (\myx(#1),0) circle;
    \draw[<-,thick,ForestGreen!25!ForestGreen,align=center]
      (\myx(#1),\yyp) -- (\myx(#1),\yyw)
      node[below,ForestGreen!40!ForestGreen] {\strut#4}; %,fill=white
    }
 
  % axis
  \draw[->,thick] (-\w*0.03,0) -- (\w*1.06,0)
                  node[right=4pt,below=6pt] {\textbf{Year}};
 
  % ticks
  \foreach \tick in {0,1,...,\n}{
    \def\x{{\tick*\w/\n}}
    \def\year{\the\numexpr \yearOne+\tick*1 \relax}
% 	\pgfmathparse{Mod(\tick-\noffset,\nskip)==0?1:0}
% 	\ifnum\pgfmathresult>0
	  \draw[thick] (\x,\lt) -- (\x,-\lt) % ten tick
	               node[below] {\textbf{\year}}; % label
% 	\else
    %   \draw[thick] (\x,\ls) -- (\x,-\ls) % ten tick
                % node[below] {$10^{\dec}$}; % label
% 	\fi
  }

  % low mass
\arrowLabelForestGreenDown(2019.1,-1.2,1.5,p-Laplacian~\cite{slepcev2019analysis})
\arrowLabelCyanUp(2019.2,1.2,2.0,ARGA~\cite{pan2019learning}\\ARVGA~\cite{pan2019learning})
\arrowLabelBlueUp(2019.35,1.2,1.0,MixHop~\cite{MixHop})
\arrowLabelBlueUp(2019.5,1.2,2.0,SGC~\cite{SGC})
\arrowLabelBlueUp(2019.7,1.2,1.0,Janossy pooling~\cite{murphy2018janossy})
\arrowLabelBlueUp(2019.9,1.2,2.0,NeuroSAT~\cite{lstmGNN})

\arrowLabelForestGreenDown(2020.1,-1.2,1.5,poisson learning~\cite{calder2020poisson})
\arrowLabelBlueUp(2020.2,1.2,2.0,DGN~\cite{DGN})
\arrowLabelBlueUp(2020.4,1.2,1.0,Scattering GCN~\cite{Scattering_GCN})
\arrowLabelBlueUp(2020.6,1.2,2.0,DropEdge~\cite{DBLP:conf/iclr/RongHXH20})
\arrowLabelBlueUp(2020.8,1.2,1.0,InfoGraph*~\cite{DBLP:conf/iclr/SunHV020})

\end{tikzpicture}
}
\caption{Chronological overview of representative GSSL methods. \textcolor{ForestGreen}{Green} indicates graph regularization methods. \textcolor{Red}{Red} indicates matrix-factorization-based methods. \textcolor{BurntOrange}{Orange} indicates random-walk-based methods. \textcolor{Cyan}{Cyan} indicates AutoEncoder-based methods. \textcolor{Blue}{Blue} indicates GNN-based methods. Models shown above the chronological axis are involved with deep learning techniques while those below the chronological axis are not. (Better viewed in color.)} 
\label{fig:chronological_gssl}
\end{figure*}
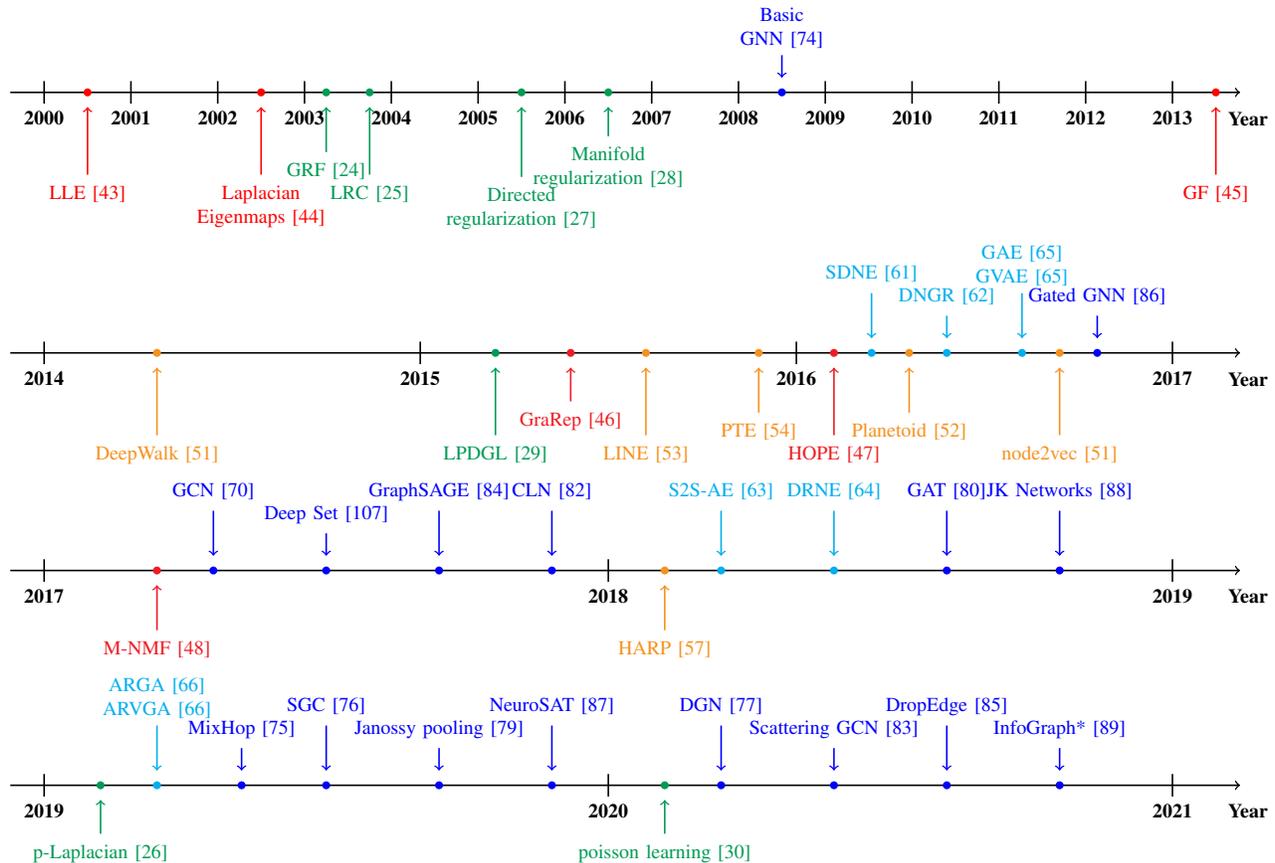

Here a chronological overview of the mentioned representative methods in this survey is provided in Figure~\ref{fig:chronological_gssl}. From 2000 to 2012, the mainstream algorithm was centered around graph regularization and matrix factorization in the early years. After the resurgence of deep learning in 2015, the field witnessed the emergence of AutoEncoder-based methods while, in the meantime, random-walk-based methods coexisted. However, with the introduction of GCN~\cite{kipf2016semi} in 2017, the GNN-based method became the dominant solution and still is a heated topic now.
\label{Chronological_appendix}

%\section{}

% \section*{Acknowledgment}

\ifCLASSOPTIONcaptionsoff
  \newpage
\fi



%\bibliographystyle{IEEEtran}
%\begin{thebibliography}{1}

% \clearpage
%\end{thebibliography}
\bibliographystyle{IEEEtran}
\bibliography{IEEEabrv,main}


%\begin{IEEEbiography}[{\includegraphics[width=1in,height=1.25in,clip,keepaspectratio]{yxlpic.jpg}}]{Xiangli~Yang}
%	Biography text here.
%\end{IEEEbiography}


%\newpage


%\vfill


\end{document}